\RequirePackage{ifluatex}
\documentclass[sn-standardnature,iicol]{sn-jnl}
\usepackage{graphicx}
\usepackage{amsmath,amssymb}
\usepackage{epstopdf}
\usepackage{bm}
\usepackage{indentfirst}
\usepackage{setspace}
\usepackage{tabulary}
\usepackage{booktabs}
\usepackage{float}
\usepackage{longtable}
\usepackage{array}
\usepackage{xcolor}
\usepackage{url}
\usepackage{multirow}
\usepackage{subcaption}
\usepackage{flushend}
\usepackage{breakurl}

\jyear{2023}

\begin{document}

\title[DCANet]{DCANet: Dual Convolutional Neural Network with Attention for Image Blind Denoising}

\author[1]{Wencong Wu}\email{cswcwu@gmail.com}

\author[1]{Guannan Lv}\email{guannanlvcn@gmail.com}

\author[1]{Yingying Duan}\email{duanyingying132@163.com}

\author[1]{Peng Liang}\email{cspengliang@126.com}

\author*[1]{Yungang Zhang}\email{yungang.zhang@ynnu.edu.cn}

\author*[1]{Yuelong Xia}\email{xyl@ynnu.edu.cn}

\affil[1]{\orgdiv{School of Information Science}, \orgname{Yunnan Normal University}, \orgaddress{\street{No. 798, Juxian Street}, \city{Kunming}, \postcode{650500}, \state{Yunnan Province}, \country{China}}}

\abstract{Noise removal of images is an essential preprocessing procedure for many computer vision tasks. Currently, many denoising models based on deep neural networks can perform well in removing the noise with known distributions (i.e. the additive Gaussian white noise). However eliminating real noise is still a very challenging task, since real-world noise often does not simply follow one single type of distribution, and the noise may spatially vary. In this paper, we present a novel dual convolutional neural network (CNN) with attention for image blind denoising, named as the DCANet. To the best of our knowledge, the proposed DCANet is the first work that integrates both the dual CNN and attention mechanism for image denoising. The DCANet is composed of a noise estimation network, a spatial and channel attention module (SCAM), and a dual CNN. The noise estimation network is utilized to estimate the spatial distribution and the noise level in an image. The noisy image and its estimated noise are combined as the input of the SCAM, and a dual CNN contains two different branches is designed to learn the complementary features to obtain the denoised image. The experimental results have verified that the proposed DCANet can suppress both synthetic and real noise effectively. The code of DCANet is available at https://github.com/WenCongWu/DCANet.}

\keywords{Image blind denoising, dual convolutional neural network, attention mechanism, noise estimation.}

\maketitle

\section{Introduction}

As one of the most significant research areas of low-level visual tasks, image denoising aims to restore clean images from noisy ones. During the past decades, many researchers have presented a number of denoising methods. Before the wide application of deep neural networks (DNNs), filtering techniques and sparse learning are widely used denoising methods. For instance, in the NLM \cite{Buades2005}, the weighted average of all pixels within the search window in an image is applied to achieve noise removal. The BM3D \cite{Dabov2007} improves the sparse representation by collaborative alteration. The trilateral weighted sparse coding (TWSC) \cite{Xu2018} accomplishes real image denoising by employing image priors. The weighted nuclear norm minimization (WNNM) \cite{Gu2014} and the multi-channel WNNM (MCWNNM) for color images \cite{Xu2017} employ the low rank approach and prior knowledge to enhance denoising performance. These denoising methods can obtain favorable denoising performance, however most of them have to contain a complex and time-consuming optimization algorithm. Meanwhile, many manually adjusted parameters are also usually required for these models to perform well, this may lead to uncertainty of their denoising performance. Therefore these models can hardly be applied in practical denoising scenes.

From the early successful deep neural networks (DNNs) based image denoising model DnCNN \cite{Zhang2017} to the present, the DNNs based denoising models have received much attention due to their superior denoising effect. However, many denoising models based on DNN are designed for the additive white Gaussian noise (AWGN) only. Moreover, in early DNN-based models, usually one model is trained only for one specific noise level, therefore it is very hard to generalize these models for other noise levels, or other types of noise \cite{Chatterjee2010}. To make the DNN based denoising models more flexible, many techniques have been proposed. For instance, a tunable noise level map is exploited as the input of the FFDNet \cite{Zhang2018}. As the map is variable or non-uniform, the FFDNet is able to handle different noise levels and the spatially varying noise can be handled as well. Nevertheless, the map of the FFDNet has to be set manually by human experience, which makes the model still far from the real denoising tasks.

For real denoising scenes, noise removal has to be completed with the unknown noise level or distribution, namely, blind denoising is required. To this end, many methods have been proposed. As an extension of the FFDNet, the CBDNet \cite{Guo2019} introduces a noise estimation sub-network, which makes the CBDNet as a fully blind denoising model. An optimal fusion function derived from the Sigal-to-Noise Ratio (SNR) was proposed in \cite{Helou2020}, and a corresponding BUIFD was then designed for image blind denoising. The AirNet model was proposed in \cite{Li2022}, which can restore low quality images corrupted by multiple degradation models by using only one single network. The TC-Net model was presented in \cite{Xue2023}, which uses the transformer combining with CNN to extract local details and global relationships. The TGVLNet model was proposed in \cite{Xiao2022}, which adopts the total generalized variation method in the network design to improve its flexibility and generalization ability. However, many existing blind denoising models have massive parameters to be learned, and their performance still can be further improved, especially in the removal of real noise.

Recently, some researchers have investigated different network structures to improve the learning ability of the denoising models. For instance, the BRDNet \cite{Tian2020} contains two different branches (the upper sub-network and the lower one), which enables the model to extract complementary features and therefore enhance the denoising performance. Pan et al. \cite{Pan2018} presented a DualCNN containing two sub-networks of different depths, where the shallow sub-network is utilized to extract the image structures, and the deeper one aims to learn the image details. Both the BRDNet and the DualCNN achieve competitive denoising performance, demonstrating that using the dual structure CNN to learn complementary features can be helpful for image denoising.

The attention mechanism can be a helpful tool for image denoising has also been verified by many researchers. For example, Anwar et al. \cite{Anwar2019} proposed the RIDNet model containing feature attention, which mainly includes four enhancement attention modules (EAM) with short skip connection and local connection. The ADNet model was developed in \cite{TianX2020}, in which an essential attention block is utilized to filter the noise information. Although both the RIDNet and ADNet achieve noticeable denoising performance, only one type of attention mechanism is used in their models, this may lead to the ignoring of the other inter-relationships of convolutional features.

Motivated by the success of the dual network structure and the attention mechanism in image denoising, in this paper, we present an effective dual CNN with attention mechanism for image blind denoising, named as the DCANet, which can achieve a competitive blind denoising performance. In our DCANet, the noise level map of a noisy image is first estimated by the noise estimation network, the noise information and the noisy image are then concatenated together as the input of the spatial and channel attention module (SCAM). The SCAM makes the model focus on the features that contain more useful contextual information. The dual CNN in our DCANet is composed of two different sub-networks, where the downsampling and dilated convolutions are used to enlarge the receptive field. The skip connections are also used to improve the training of the network. The DCANet owns the following characteristics:

(1) As far as we know, the proposed DCANet is the first model that investigates the integration of dual CNN and attention mechanism for image blind denoising.

(2) A novel dual CNN denoising network and a new noise estimator are designed in our denoising model.

(3) The DCANet is capable of obtaining competitive denoising results compared with other state-of-the-art denoising models. Moreover, the DCANet has a simpler model structure compared with many existing blind denoising models.

The remainder of this paper is organized as follows. Section \ref{Related_work} reviews the related denoising methods. Our proposed DCANet model is introduced in Section \ref{Proposed_model}. Section \ref{Experiment} reports the experimental results. The paper is summarized in Section \ref{Conclusion}.

\section{Related work}\label{Related_work}
\subsection{Image blind denoising models}
Some early DNN based denoising models have already made the effort to blind denoising. For example, the famous DnCNN \cite{Zhang2017} can be used for real denoising, however its performance is far from satisfaction. Zhang et al. \cite{ZhangZGZ2017} proposed the IRCNN model by learning deep denoiser prior to realize blind denoising, and promising results were reported. Peng et al. \cite{Peng2019} designed a dilated residual network for blind denoising, where the authors claimed that dilated convolution and skip connection can be helpful for noise removal.

To capture the noise feature better, many researchers have tried to utilize a noise estimation network to estimate the noise distributions and levels in an image. Based on the DnCNN, the BUIFD model with a noise estimator was proposed in \cite{Helou2020}. The BUIFD can handle unknown and universal noise by a single model. Soh et al. \cite{Soh2020} presented a deep universal blind denoising (DUBD) network, where a conditional estimation network (CENet) is introduced for incorporating human prior knowledge, and a tunable denoising sub-network is designed for noise removal. Guo et al. \cite{Guo2019} designed the CBDNet model for blind denoising, where a noise estimation sub-network is used to estimate the noise level, and asymmetric learning is adopted to restrain the possible under-estimation of the noise level. Yue et al. \cite{Yue2019} presented a novel variational denoising network (VDN) for image blind denoising. The VDN contains a denoising network and a Sigma network, the Sigma network is applied to estimate the noise distribution in a noisy image. Kim et al. \cite{Kim2020} proposed an adaptive instance normalization denoising network (AINDNet), and the adaptive instance normalization (AIN) is developed to regularize the network, so that the model will avoid overfitting to the synthetical noisy images, and a noise level estimator is applied to estimate the noise level.

It can be found that currently most of the blind denoising models utilize a noise estimation sub-network as the solution for achieving blind denoising, and the effectiveness of this solution has been verified. In our proposed denoising model, a noise estimation block is also developed. Compared with the noise estimation networks of the CBDNet \cite{Guo2019}, VDN \cite{Yue2019}, and DUBD \cite{Soh2020}, our noise estimator is designed to contain more convolutional layers, which we think can help to extract more complex noise information.

\subsection{Attention-guided denoising models}
Extracting and selecting appropriate features is extremely important for image processing tasks\cite{Du2019, Li2021, Liang2021}. However, for images with complex textures, it is very difficult to obtain features with high discriminative ability \cite{Li2019}. To solve this problem, the attention mechanism is proposed in \cite{Larochelle2010}, which is able to focus on the salient regions in images, and can capture the useful features better. Since then the attention mechanisms have been widely utilized in various image analysis tasks, including image denoising.

A residual dilated attention network (RDAN) was designed in \cite{Hou2019} for blind denoising, which contains two attention blocks, the attention mechanism in RDAN enhances the restoration of texture details. A multi-stage denoising architecture MPRNet \cite{Zamir2021} was developed for image restoration, where the supervised attention modules and channel attention blocks were introduced and used to promote the restoration effect. A very deep residual non-local attention network (RNAN) was proposed in \cite{Zhang2019} for image restoration, the attention-based RNAN achieves competitive performances on different image restoration tasks, including image denoising. The channel and space attention neural network (CSANN) was presented in \cite{Wang2021}, the CSANN can predict the residual image to eliminate noise. An adaptive consistency prior (ACP) was presented in DeamNet \cite{Ren2021}, in which the dual element-wise attention mechanism (DEAM) modules were designed and utilized to enhance its learning ability. A multi-attention complementary fusion network (MACFNet) \cite{Yu2022} was designed for image denoising, where multiple dimensional attention mechanisms were merged to enhance denoising performance and details restoration.

It can be observed that the MPRNet \cite{Zamir2021}, RNAN \cite{Zhang2019}, CSANN \cite{Wang2021}, and MACFNet \cite{Yu2022} all use multiple attention mechanisms to exploit different dependencies in convolutional features, and remarkable denoising results were produced by these models. The success of these attention-guided models reveals that the attention mechanism can be an effective tool for image denoising.

\subsection{Dual convolutional neural network for image denoising}
In general, very deep networks may cause the vanishing or exploding of gradients, which affects the performance of the network. To address this issue, some researchers have tried to promote the learning ability of the denoising model by increasing the width of the network instead of the depth. The dual CNN adopts such network composition strategy, where two parallel branches (sub-networks) are contained in a dual CNN, and usually different network structures are designed for the two branches to boost the learning ability.

The dual CNN structures for image denoising tasks have also been investigated by some researchers. Tian et al. \cite{Tian2020} presented a BRDNet model, which consists of a dual CNN containing two different sub-networks. As different network architectures can capture different image features \cite{ZhangT2018}, the BRDNet therefore can extract complementary features to improve its denoising performance, impressive denoising results were reported by BRDNet. Later, Tian et al. \cite{Tian2021} designed another dual denoising network (DudeNet) with two different branches to achieve effective noise removal, and the model also obtains favorable results. For general low-level vision tasks, Pan et al. \cite{Pan2022} presented a dual CNN (DualCNN). The DualCNN is composed of two different parallel branches, where one branch is used for capturing the image structures, and another branch is applied to extract the image details.

\section{The proposed model}\label{Proposed_model}
\subsection{Network architecture}
\begin{figure*}[htbp]
	\begin{center}
		\includegraphics[width=\textwidth]{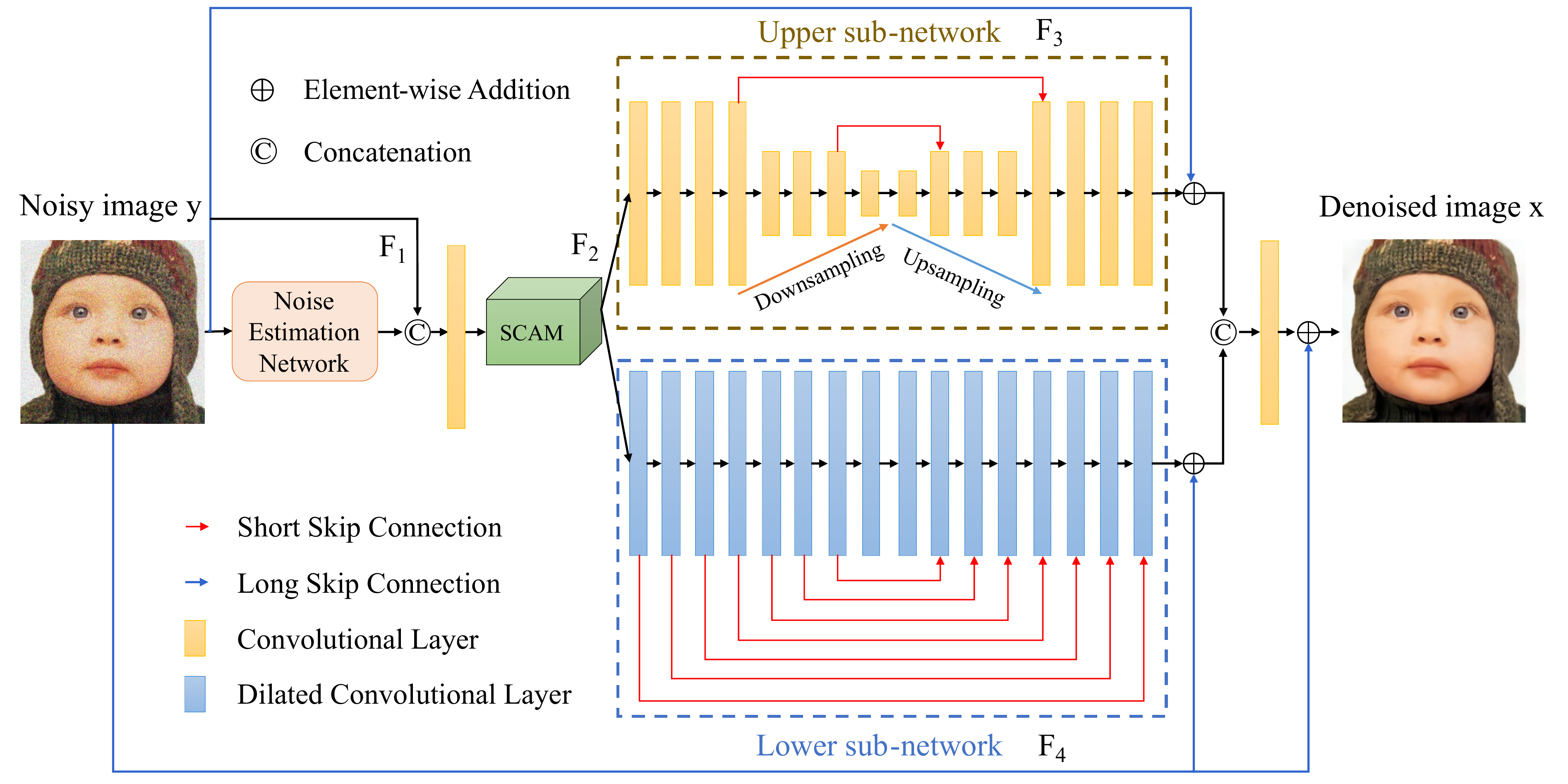}
		\caption{The network structure of the proposed model.}
		\label{fig:DCANet}
	\end{center}
\end{figure*}

In this section, we introduce the proposed DCANet. Fig. \ref{fig:DCANet} shows the architecture of the DCANet. The DCANet model consists of a noise estimation network, a spatial and channel attention module (SCAM), and a dual CNN. The denoising process of the DCANet can be formulated as follows:

\begin{equation}
    x = f_{DCANet}(y),
\end{equation}

\noindent where $x$ and $y$ denote the denoised and noisy image, respectively. The noise estimation network is utilized to estimate the noisy information of the noisy image, as formulated below:

\begin{equation}
\label{noise_feature}
    F_{1} = Net_{est}(y),
\end{equation}

\noindent where $Net_{est}$ represents the noise estimation network, and $F_{1}$ is the predicted noise information. Then a convolution layer takes the concatenation of the noisy image and the estimated noise level map as its input, producing a feature map with 64 channels, which is sent to the SCAM for feature filtering. The focused feature $F_{2}$ is extracted by the SCAM, as formulated below:

\begin{equation}
    \begin{aligned}
        F_{1}^{1} &= K_{s} * [F_{1}, y],\\
        F_{2} &= SCAM(F_{1}^{1}),
    \end{aligned}
\end{equation}

\noindent where $[ , ]$ denotes the concatenation operation, $K_{s}$ is the standard convolutional kernel, and $*$ represents convolution. The dual CNN is then applied to capture complementary features from $F_2$, as formulated below:

\begin{equation}
    \begin{aligned}
        F_{3} &= USN(F_{2}),\\
        F_{4} &= LSN(F_{2}),
    \end{aligned}
\end{equation}

\noindent where $USN$ and $LSN$ represent the upper and lower sub-networks, respectively. $F_{3}$ and $F_{4}$ denote the extracted features. Finally, the denoised image $x$ is obtained by concatenating the two sub-networks, as formulated below:

\begin{equation}
    x = K_{s} * [F_{3} + y, F_{4} + y] + y.
\end{equation}

\noindent In our model, the channel number is set to 64, which can balance network complexity and denoising performance. The BN and ReLU layers in each layer of convolution or dilated convolution are not presented in Fig. \ref{fig:DCANet}. Although there are many existing methods utilize the dual CNN structures or incorporate attention mechanisms for image denoising, as far as we know, no previous work has reported the combination of the dual CNN and attention mechanism for image blind denoising. In the following subsections, we introduce the noise estimator, the attention module, and the dual CNN in detail.

% The network architecture of the proposed DCANet

\subsubsection{Noise estimation network}
\begin{figure*}[htbp]
	\centering
	\begin{subfigure}{0.49\linewidth}
		\centering
		\includegraphics[width=0.99\linewidth]{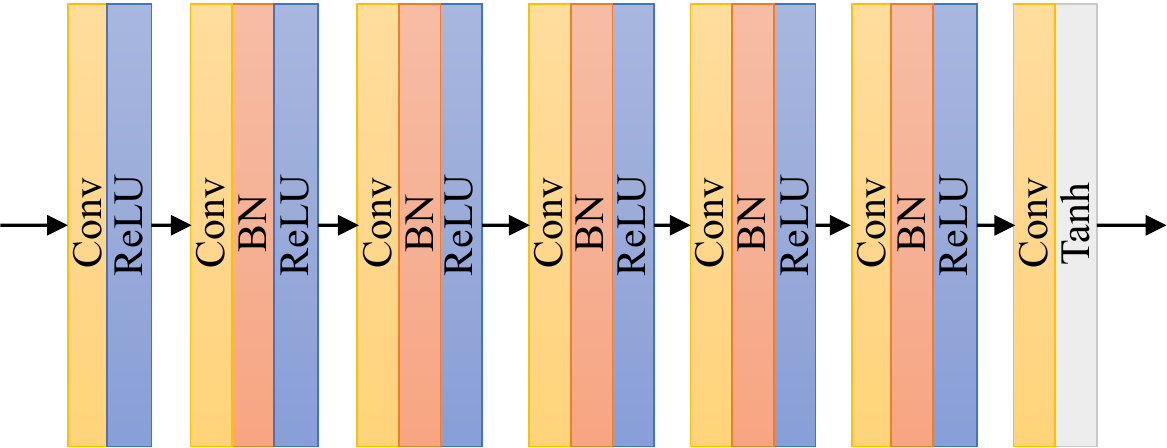}
		\caption{The noise estimation network of our DCANet.}
	\end{subfigure}\vspace{2mm}
    \centering
	\begin{subfigure}{0.49\linewidth}
		\centering
		\includegraphics[width=0.60\linewidth]{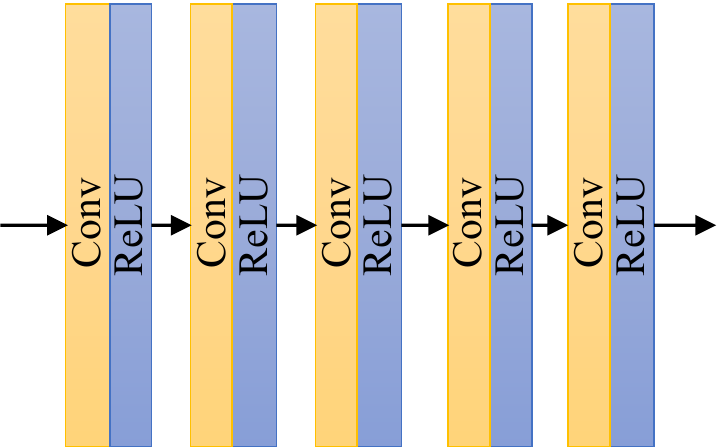}
		\caption{The noise estimation network of CBDNet \cite{Guo2019}.}
	\end{subfigure}\vspace{2mm}
    \centering
	\begin{subfigure}{0.49\linewidth}
		\centering
		\includegraphics[width=0.725\linewidth]{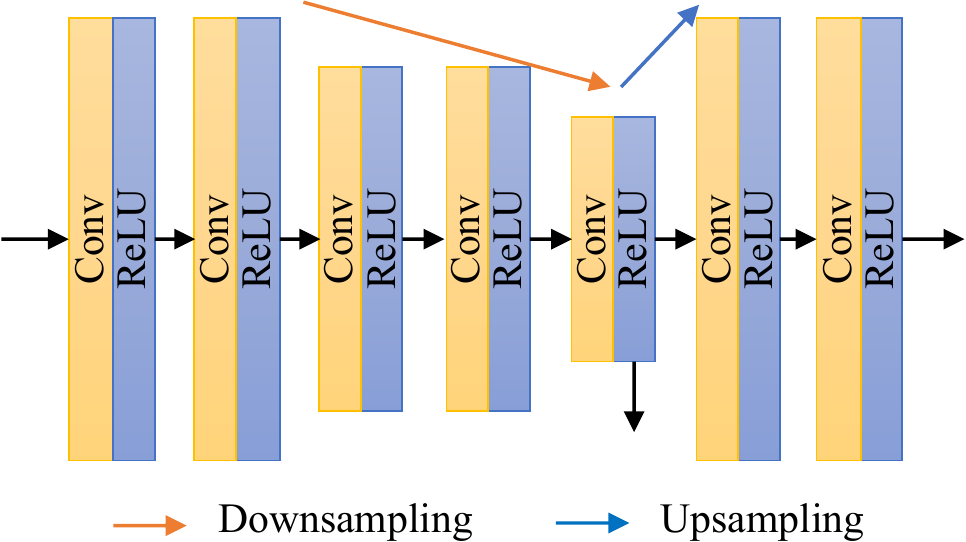}
		\caption{The noise estimation network of AINDNet \cite{Kim2020}.}
	\end{subfigure}
    \centering
	\begin{subfigure}{0.49\linewidth}
		\centering
		\includegraphics[width=0.60\linewidth]{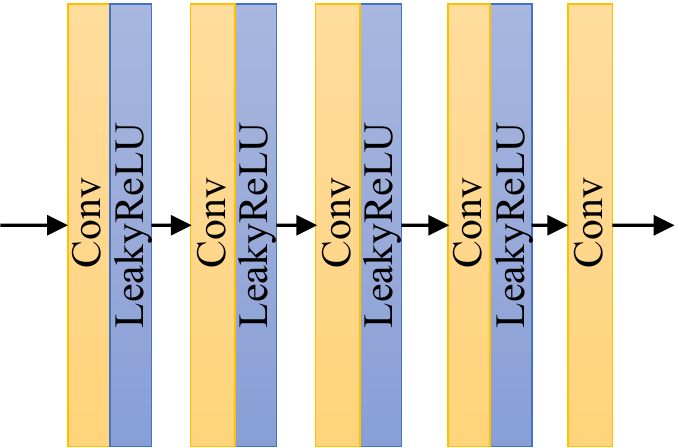}
		\caption{The noise estimation network of VDN \cite{Yue2019}.}
	\end{subfigure}
\caption{The noise estimation network architectures of our DCANet (a), CBDNet (b), AINDNet (c), and VDN (d).}
\label{fig:four_est}
\end{figure*}

To accurately estimate the noise in an image, unlike the noise estimation network of CBDNet \cite{Guo2019} and VDN \cite{Yue2019}, our noise estimation network contains more convolutional layers, which can make the network extract complex noise information. The architectures of noise estimation sub-networks of the proposed DCANet, CBDNet \cite{Guo2019}, AINDNet \cite{Kim2020}, and VDN \cite{Yue2019} can be seen in Fig. \ref{fig:four_est}. Our estimator has 7 convolutional layers, where the Convolution (Conv), Rectified Linear Units (ReLU) \cite{Krizhevsky2012}, Batch Normalization (BN) \cite{Ioffe2015} and Tanh \cite{Malfliet1996} are used in the network. Specifically, ``Conv+ReLU'', ``Conv+BN+ReLU'' and ``Conv+Tanh'' are used in the first, the middle, and the last layers respectively. The noise feature $F_1$ in Eqn. (\ref{noise_feature}) can be obtained by:

\begin{equation}
    \begin{aligned}
        F_{0}^{0} &= \phi(K_{s} * y),\\
        F_{0}^{i+1} &= \phi(BN(K_{s} * F_{0}^{i})), i \in \{0, 1, 2, 3, 4\},\\
        F_{1} &= \xi(K_{s} * F_{0}^{5}),
    \end{aligned}
\end{equation}

\noindent where $y$, $K_{s}$, and $F_{1}$ are the noisy image, standard convolutional kernel, and the estimated noise level, respectively. $\phi$ and $\xi$ represent the ReLU and Tanh activation functions, respectively. $F_{0}^{0}$ is the feature obtained by the first layer, and $F_{0}^{i+1}$ denotes the features sequentially extracted by the five middle convolutional layers, and $F_{1}$ is the noise feature obtained by the last layer of the noise estimator. The zero padding is used in our noise estimator to keep the feature map size constant.

It should be noted that both our noise estimation network and the estimator in AINDNet \cite{Kim2020} have seven convolutional layers. Different from ours, the noise estimation network of the AINDNet has two downsampling operations. Although a larger receptive field size is beneficial to extract more image information, the downsampling operations yet will bring information loss. In particular, compared with the noise estimation sub-networks of the CBDNet, VDN, and AINDNet, our estimation network incorporates multiple BN layers, which not only accelerate network training, but also enhance network performance.

% The four noise estimation network architectures

\subsubsection{Spatial and channel attention module}
\begin{figure*}[htbp]
	\begin{center}
		\includegraphics[width=0.9\textwidth]{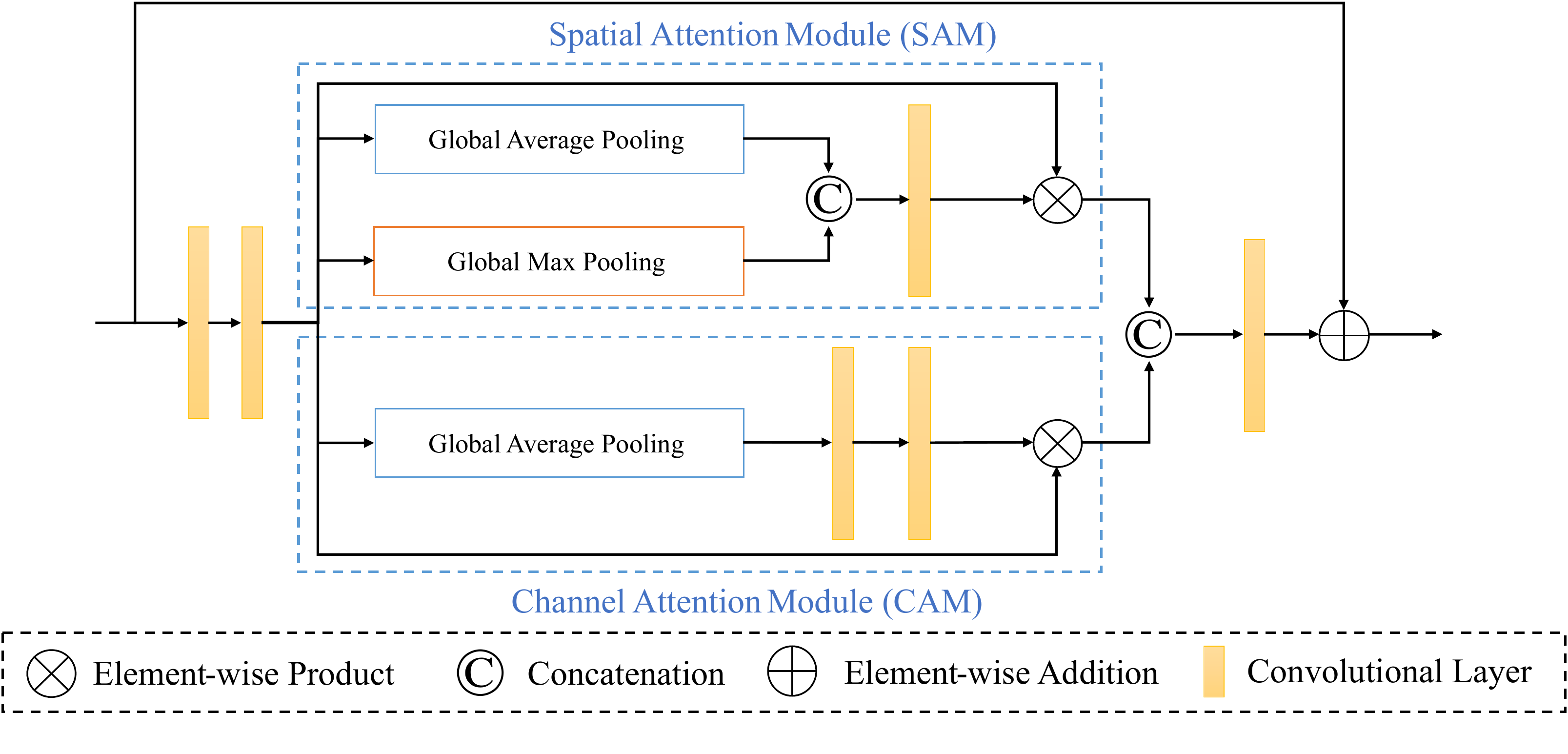}
		\caption{The structure of the SCAM.}
		\label{fig:SCAM}
	\end{center}
\end{figure*}

To leverage the useful features in images, many denoising models employ attention mechanisms to improve their learning ability, such as the ADNet \cite{TianX2020}, RDAN \cite{Hou2019}, RNAN \cite{Zhang2019}, CSANN \cite{Wang2021}, and DeamNet \cite{Ren2021}. Following the attention network introduced in CSANN \cite{Wang2021}, our proposed DCANet also utilizes the spatial and channel attention for feature selection, which is presented in Fig. \ref{fig:SCAM}. Our spatial and channel attention module (SCAM) consists of the parallel spatial attention module (SAM) and channel attention module (CAM), which contains the following operations: Conv, PReLU \cite{He2015}, Global Average Pooling (GAP) \cite{Hu2018}, Global Max Pooling (GMP) \cite{Woo2018}, BN, ReLU, Sigmoid \cite{Han1995}. The implemental procedure of the SCAM can be expressed as follows:

\begin{equation}
    \begin{aligned}
        F_{1}^{2} &= \varphi(K_{s} * F_{1}^{1}),\\
        F_{1}^{3} &= K_{s} * F_{1}^{2},\\
        F_{1}^{4} &= \delta(\phi(BN(K_{s} * [GAP(F_{1}^{3}), GMP(F_{1}^{3})]))) \cdot F_{1}^{3},\\
        F_{1}^{5} &= \delta(K_{s} * \phi(K_{s} * GAP(F_{1}^{3}))) \cdot F_{1}^{3},\\
        F_{2} &= K_{s} * [F_{1}^{4}, F_{1}^{5}] + F_{1}^{1},
    \end{aligned}
\end{equation}

\noindent where $F_{1}^{i}$ ($i \in$ \{1, 2, 3, 4, 5\}) and $F_{2}$ denote the captured features. $\varphi$, $\delta$, and $\phi$ represent PReLU, Sigmod, and ReLU activation functions, respectively. The '$\cdot$' and $[ , ]$ are the element-wise product and concatenation operation. The SCAM is employed to exploit the inter-spatial and inter-channel relationships of the convolutional features. The informative features are retained by the SCAM module, and the unimportant ones are suppressed.

% The architecture of the proposed SCAM

\subsubsection{Dual convolutional neural network}
It has been verified that expanding the width of the denoising network is also a useful way to improve network performance \cite{Szegedy2015}. In this work, we develop a dual CNN with two different sub-networks for image denoising, which can be seen in the brown and blue dotted boxes in Fig. \ref{fig:DCANet}, where the upper sub-network is a U-shaped network, and the lower branch is a dilated convolution network. It has been verified that using different sub-networks in dual CNN can enhance the performance of the whole network, as the complementary features can be learned by the two different branches \cite{Tian2020, ZhangT2018, Tian2021}. The upper branch of our dual CNN involves the following operations: Conv, ReLU, BN, max-pooling (Downsampling) \cite{Ranzato2007}, bilinear interpolation (Upsampling) \cite{Huang2020}, and skip connections \cite{He2016, Huang2017}. The lower branch includes Dilated Convolution (DConv) \cite{Yu2015}, ReLU, BN, and the skip connections as well. The outputs of two different branches are concatenated through a concatenation operation.

Compared with the existing dual CNN based denoising models such as the BRDNet \cite{Tian2020} and DualCNN \cite{Pan2022}, our proposed dual network is designed to own a larger receptive field, which allows the network to capture more contextual information. Specifically, in our dual network, the upper branch employs two times of downsampling to increase the receptive field. However, the downsampling operation may cause the loss of image information. Therefore, two skip connections are applied in the upper sub-networks to suppress the loss. The upsampling operations are employed to recover the size of the feature map.

The lower sub-network applies the dilated convolution to increase the size of the receptive field, and the symmetric skip connections are utilized to speed up network training. Particularly, we adopt the hybrid dilated convolution (HDC) in the lower branch, it has been verified that the HDC can remove the gridding phenomenon effectively and improve network performance \cite{Yu2017, Wang2018}. The dilated rate of each dilated convolution layer in the lower branch is shown in Table \ref{tab:rate}.

\begin{table*}[htbp]
\centering
\caption{The dilated rates of different layers in the lower sub-network.}
\label{tab:rate}
\begin{tabular}{ccccccccccccccccc}
\cline{1-17}
Layer & 1 & 2 & 3 & 4 & 5 & 6 & 7 & 8 & 9 & 10 & 11 & 12 & 13 & 14 & 15 & 16 \\
\cline{1-17}
Dilated rate & 1 & 2 & 3 & 4 & 5 & 6 & 7 & 8 & 7 & 6 & 5 & 4 & 3 & 2 & 1 & 1\\
\cline{1-17}
\end{tabular}
\end{table*}

\subsection{Loss functions}
Mean squared error (MSE) is a widely used loss function for optimizing neural networks \cite{Yang2019, ZhangP2018}. For the AWGN removal, we optimize our DCANet by the following loss function:
\begin{equation}
\begin{aligned}
\mathcal{L} &= \frac{1}{2N} \sum_{i=1}^N \left\|f(y_i, \theta)- x_i\right\|^2\\
            &= \frac{1}{2N} \sum_{i=1}^N \left\|\hat{x}_i - x_i\right\|^2,
\end{aligned}
\end{equation}
\noindent where $x_i$, $y_i$, and $\hat{x}_i$ are the clean, noisy, and denoised images, respectively. $N$ is the number of clean-noisy image patches, and $\theta$ represents the parameters of the DCANet.

For removing the unknown real noise, as the high-frequency textures will be lost due to the square penalty, then the MSE often produces a blurred and overly smoothed visual effect. Therefore, we select the Charbonnier loss \cite{Lai2017} as the reconstruction loss to optimize our model for real noise removal. Moreover, to preserve the fidelity and authenticity of the high-frequency details, we utilize the edge loss proposed in \cite{Jiang2020} to constrain the loss of the high-frequency components between the ground-truth image $x$ and the denoised image $\hat{x}$.

Our proposed DCANet contains a noise estimator to predict the noise level $\sigma(y)$ in a noisy image $y$. For real noise removal, the estimated noise level should be avoided to be too smooth, especially for the spatially variant noise, which may change dramatically throughout the whole noisy image. Therefore, we employ a total variation (TV) regularizer \cite{Guo2019} to restrict the smoothness of the estimated noise level. Therefore, the total loss of our DCANet is expressed as:
\begin{equation}
\mathcal{L} = \mathcal{L}_{char}(\hat{x}, x) + \lambda_{edge}\mathcal{L}_{edge}(\hat{x}, x) + \lambda_{TV}\mathcal{L}_{TV}(\sigma(y)),
\label{eq.2}
\end{equation}
where we empirically set $\lambda_{edge}$ and $\lambda_{TV}$ to 0.1 and 0.05 respectively, and $\mathcal{L}_{char}$ is the Charbonnier loss, which is represented as:
\begin{equation}
\mathcal{L}_{char} = \sqrt{\left\|\hat{x} - x\right\|^2 + \epsilon^2},
\end{equation}
where the constant $\epsilon$ is set as $10^{-3}$. $\mathcal{L}_{edge}$ is the edge loss, which can be expressed as:
\begin{equation}
\mathcal{L}_{edge} = \sqrt{\left\|\bigtriangleup{(\hat{x})} - \bigtriangleup{(x)}\right\|^2 + \epsilon^2},
\end{equation}
where $\bigtriangleup$ denotes the Laplacian operator \cite{Kamgar1999}. $\mathcal{L}_{TV}$ is defined as:
\begin{equation}
\mathcal{L}_{TV} = \left\|\bigtriangledown_{h}{\sigma(y)}\right\|_{2}^{2} + \left\|\bigtriangledown_{v}{\sigma(y)}\right\|_{2}^{2},
\end{equation}
where $\bigtriangledown_{h} (\bigtriangledown_{v})$ is the gradient operator along the horizontal (vertical) direction.

\section{Experiments and results}\label{Experiment}
\subsection{Datasets}
The DIV2K dataset \cite{Agustsson2017} containing 800 high-resolution color images was used to train the proposed DCANet for the AWGN removal, and we resized these high-resolution images into the images of the size $512 \times 512$. Moreover, the images were grayscaled to train the DCANet for grayscale image denoising. According to the size of the receptive field of the DCANet, the training images were randomly cropped into patches of the size $140 \times 140$. The AWGN within a noise level range of $[0, 75]$ is added to the clean image patches to generate their noisy counterparts. To augment the training samples, rotation and flipping were utilized. We tested our model on five public commonly used datasets, including Set12 \cite{Roth2005}, BSD68 \cite{Roth2005}, CBSD68 \cite{Roth2005}, Kodak24 \cite{Kodak24} and McMaster \cite{Zhang2011}.

For real noise removal, we select the SIDD training data \cite{Abdelhamed2018} and RENOIR dataset \cite{Anaya2018} for model training. The SIDD training data and the RENOIR dataset consist of 320 and 240 pairs of noisy images and the near noise-free counterparts, respectively. We randomly cut the images in these datasets into image patches of size $140 \times 140$, and applied rotation and flipping operations for data augmentation. The SIDD validation set \cite{Abdelhamed2018} and DND sRGB dataset \cite{Plotz2017} were used for model evaluation.

\subsection{Experimental settings}
All our experiments were performed on a PC equipped with a CPU of Intel(R) Core(TM) i7-11700KF, 32 GB RAM, and a GPU of NVIDIA GeForce RTX 3080Ti. The proposed DCANet was trained for grayscale and color images, respectively. The training of the DCANet for the synthetic noise cost about 48 hours. For real image denoising, it took about 45 hours to train the model.

The network parameters of the DCANet were optimized by the Adam optimizer \cite{Kingma2014}. For the AWGN removal, the DCANet was trained for 600,000 iterations, during which the initial learning rate is $10^{-4}$ and then decreases by half every 100,000 iterations, and the batch size was set to 24. For real image denoising, we used 120 epochs to train the model, during which the initial learning rate was $2\times10^{-4}$ and steadily reduced to $10^{-6}$ using cosine annealing strategy \cite{Loshchilov2017}, the batch size was set to 16. For other hyper-parameters of the Adam algorithm, we used the default settings.

\subsection{Ablation study}
To verify the effectiveness of the proposed denoising architecture, especially the effect of SCAM, dual CNN, and skip connections, we trained ten different networks for grayscale image denoising. The ten different models and their corresponding performance are shown in Table \ref{tab:Set12_val}. The Set12 \cite{Roth2005} dataset was utilized for our ablation evaluation, and the noise level was set to 75.

In Table \ref{tab:Set12_val}, one can find that short and long skip connections can improve the network performance. Besides, the denoising performance of the models using only a single attention mechanism (SAM or CAM) is better than that of the model using serial attention mechanisms, but lower than that of the model using parallel attention mechanisms (SCAM in the proposed model), which shows the effectiveness of the proposed SCAM. In addition, the denoising performance of the model using a single sub-network is inferior to the dual CNN structure. The results validate that using SCAM, dual CNN, and skip connections can improve the model performance.

%PSNR and SSIM results of Set12 at noise level 75
\begin{table*}[htbp]
\centering
\caption{Quantitative comparison results of the ten models containing different components on the Set12 dataset. The noise level is set to 75. The best results are bolded.}
\label{tab:Set12_val}
\begin{tabular}{cccc}
\cline{1-4}
Model ID & Models & PSNR & SSIM\\
\cline{1-4}
1 & DCANet without short skip connection & 25.60 & 0.7356\\
\cline{1-4}
2 & DCANet without long skip connection & 25.72 & 0.7447\\
\cline{1-4}
3 & DCANet without SCAM & 25.65 & 0.7406\\
\cline{1-4}
4 & DCANet without SAM & 25.73 & 0.7453\\
\cline{1-4}
5 & DCANet without CAM & 25.71 & 0.7444\\
\cline{1-4}
6 & DCANet with serial attention mechanism ( SAM(front) and CAM (behind) ) & 25.62 & 0.7377\\
\cline{1-4}
7 & DCANet with serial attention mechanism ( CAM(front) and SAM (behind) ) & 25.66 & 0.7409\\
\cline{1-4}
8 & DCANet with only upper sub-network & 25.52 & 0.7325\\
\cline{1-4}
9 & DCANet with only lower sub-network & 25.70 & 0.7443\\
\cline{1-4}
10 & Proposed DCANet model & $\mathbf{25.75}$ & $\mathbf{0.7458}$\\
\cline{1-4}
\end{tabular}
\end{table*}

\subsection{The additive white Gaussian noise removal}
In this subsection, we show the results of our DCANet on grayscale and color images corrupted by the AWGN. The BM3D \cite{Dabov2007}, WNNM \cite{Gu2014}, MCWNNM \cite{Xu2017}, TNRD \cite{Chen2017}, DnCNN \cite{Zhang2017}, BUIFD \cite{Helou2020}, IRCNN \cite{ZhangZGZ2017}, FFDNet \cite{Zhang2018}, BRDNet \cite{Tian2020}, ADNet \cite{TianX2020}, DudeNet \cite{Tian2021}, DSNetB \cite{Peng2019}, RIDNet \cite{Anwar2019}, AINDNet \cite{Kim2020} and AirNet \cite{Li2022} were used to compare with our DCANet.

Table \ref{tab:Set12_PSNR} lists the PSNR values at different noise levels of the compared denoising models on the Set12 dataset. One can see that the denoising performances of the DCANet are slightly behind the BRDNet and ADNet at noise levels 15 and 25, however our DACNet beats other methods at noise level 50. It also should be noted that for the image ``Barbara", the traditional methods BM3D and WNNM obtain significant denoising results, we think the reason is that this image contains rich repetitive structures, and the non-local self-similarity learning based methods can capture such structures better.

%PSNR results of Set12 at different noise levels
\begin{table*}[htbp]\tiny
\centering
\caption{Quantitative comparison results on the Set12 dataset. The three best results are respectively emphasized in red, blue and green.}
\label{tab:Set12_PSNR}
\begin{tabular}{p{35pt}<{\centering}p{43pt}<{\centering}p{15pt}<{\centering}p{20pt}<{\centering}p{20pt}<{\centering}p{20pt}<{\centering}p{20pt}<{\centering}
p{15pt}<{\centering}p{15pt}<{\centering}p{15pt}<{\centering}p{20pt}<{\centering}p{15pt}<{\centering}p{15pt}<{\centering}p{20pt}<{\centering}p{20pt}<{\centering}}
\cline{1-15}
Noise levels & Models & C.man & House & Peppers & Starfish &  Monar. &  Airpl. & Parrot &  Lena &  Barbara &  Boat &  Man & Couple & Average\\
\cline{1-15}
\multirow{22}*{$\sigma$=15} & BM3D \cite{Dabov2007} & 31.91 & 34.93 & 32.69 & 31.14	& 31.85	& 31.07	& 31.37	& 34.26	& \textcolor{blue}{33.10}	& 32.13	& 31.92	& 31.10	& 32.37\\
\cline{2-15}
    & WNNM \cite{Gu2014} & 32.17	& \textcolor{green}{35.13}	& 32.99	& 31.82 & 32.71	& 31.39	& 31.62	& 34.27	& \textcolor{red}{33.60}	& 32.27	& 32.11	& 32.17	& 32.70\\
\cline{2-15}
    & TNRD \cite{Chen2017} & 32.19	& 34.53	& 33.04	& 31.75	& 32.56	& 31.46	& 31.63	& 34.24	& 32.13	& 32.14	& 32.23	& 32.11	& 32.50\\
\cline{2-15}
    & DnCNN-S \cite{Zhang2017} & 32.61 & 34.97 & 33.30 & \textcolor{green}{32.20} & 33.09 & 31.70 & 31.83 & 34.62 & 32.64 & 32.42 & \textcolor{green}{32.46} & 32.47 & 32.86\\
\cline{2-15}
    & IRCNN \cite{ZhangZGZ2017} & 32.55 & 34.89 & 33.31	& 32.02	& 32.82	& 31.70	& 31.84	& 34.53	& 32.43	& 32.34	& 32.40	& 32.40	& 32.77\\
\cline{2-15}
    & BUIFD \cite{Helou2020} & 31.74	& 34.78	& 32.80	& 31.92	& 32.77	& 31.34 & 31.39	& 34.38	& 31.68	& 32.18	& 32.25	& 32.22	& 32.46\\
\cline{2-15}
    & FFDNet \cite{Zhang2018} & 32.43	& 35.07	& 33.25	& 31.99	& 32.66	& 31.57	& 31.81	& 34.62	& 32.54	& 32.38	& 32.41	& 32.46	& 32.77\\
\cline{2-15}
    & BRDNet \cite{Tian2020} & \textcolor{blue}{32.80} & \textcolor{red}{35.27} & \textcolor{blue}{33.47} & \textcolor{blue}{32.24} & \textcolor{red}{33.35} & \textcolor{blue}{31.85} & \textcolor{red}{32.00} & \textcolor{red}{34.75} & \textcolor{green}{32.93} & \textcolor{blue}{32.55} & \textcolor{red}{32.50} & \textcolor{red}{32.62} & \textcolor{red}{33.03} \\
\cline{2-15}
    & ADNet \cite{TianX2020} & \textcolor{red}{32.81} & \textcolor{blue}{35.22} & \textcolor{red}{33.49} & 32.17 & \textcolor{green}{33.17} & \textcolor{red}{31.86} & \textcolor{blue}{31.96} & \textcolor{blue}{34.71} & 32.80 & \textcolor{red}{32.57} & \textcolor{blue}{32.47} & \textcolor{blue}{32.58} & \textcolor{blue}{32.98}\\
\cline{2-15}
    & DudeNet \cite{Tian2021} & \textcolor{green}{32.71} & \textcolor{green}{35.13} & \textcolor{green}{33.38} & \textcolor{red}{32.29} & \textcolor{blue}{33.28} & \textcolor{green}{31.78} & \textcolor{green}{31.93} & 34.66 & 32.73 & \textcolor{green}{32.46} & \textcolor{green}{32.46} & \textcolor{green}{32.49} & \textcolor{green}{32.94}\\
\cline{2-15}
    & DCANet & 32.43	& \textcolor{green}{35.13}	& 33.09	& 32.06	& 33.16	& 31.55	& 31.80	& \textcolor{green}{34.70}	& 32.76	& 32.42	& 32.40	& 32.48	& 32.83\\
\cline{1-15}
\multirow{22}*{$\sigma$=25} & BM3D \cite{Dabov2007} & 29.45 & 32.85 & 30.16 & 28.56 & 29.25 & 28.42 & 28.93 & 32.07 & \textcolor{blue}{30.71} & 29.90 & 29.61 & 29.71 & 29.97 \\
\cline{2-15}
    & WNNM \cite{Gu2014} & 29.64 & 33.22 & 30.42 & 29.03 & 29.84 & 28.69 & 29.15 & 32.24 & \textcolor{red}{31.24} & 30.03 & 29.76 & 29.82 & 30.26 \\
\cline{2-15}
    & TNRD \cite{Chen2017} & 29.72 & 32.53 & 30.57 & 29.02 & 29.85 & 28.88 & 29.18 & 32.00 & 29.41 & 29.91 & 29.87 & 29.71 & 30.06\\
\cline{2-15}
    & DnCNN-S \cite{Zhang2017} & 30.18 & 33.06 & 30.87 & 29.41 & 30.28 & 29.13 & 29.43 & 32.44 & 30.00 & 30.21 & \textcolor{green}{30.10} & 30.12 & 30.43 \\
\cline{2-15}
    & IRCNN \cite{ZhangZGZ2017} & 30.08 & 33.06 & 30.88 & 29.27 & 30.09 & 29.12 & 29.47 & 32.43 & 29.92 & 30.17 & 30.04 & 30.08 & 30.38 \\
\cline{2-15}
    & BUIFD \cite{Helou2020} & 29.42	& 33.03	& 30.48	& 29.21	& 30.20	& 28.99	& 28.94	& 32.20	& 29.18	& 29.97	& 29.88	& 29.90	& 30.12\\
\cline{2-15}
    & FFDNet \cite{Zhang2018} & 30.10 & \textcolor{green}{33.28} & 30.93 & 29.32 & 30.08 & 29.04 & 29.44 & 32.57 & 30.01 & 30.25 & \textcolor{blue}{30.11} & 30.20 & 30.44\\
\cline{2-15}
    & BRDNet \cite{Tian2020} & \textcolor{red}{31.39} & \textcolor{red}{33.41} & \textcolor{blue}{31.04} & \textcolor{blue}{29.46} & \textcolor{blue}{30.50} & \textcolor{red}{29.20} & \textcolor{red}{29.55} & \textcolor{blue}{32.65} & 30.34 & \textcolor{green}{30.33} & \textcolor{red}{30.14} & \textcolor{red}{30.28} & \textcolor{red}{30.61}\\
\cline{2-15}
    & ADNet \cite{TianX2020} & \textcolor{blue}{30.34} & \textcolor{red}{33.41} & \textcolor{red}{31.14} & 29.41 & 30.39 & \textcolor{blue}{29.17} & \textcolor{blue}{29.49} & \textcolor{green}{32.61} & 30.25 & \textcolor{red}{30.37} & 30.08 & \textcolor{green}{30.24} & \textcolor{blue}{30.58}\\
\cline{2-15}
    & DudeNet \cite{Tian2021} & \textcolor{green}{30.23} & 33.24 & \textcolor{green}{30.98} & \textcolor{red}{29.53} & \textcolor{green}{30.44} & \textcolor{green}{29.14} & \textcolor{green}{29.48} & 32.52 & 30.15 & 30.24 & 30.08 & 30.15 & 30.52\\
\cline{2-15}
    & DCANet & 30.18	& \textcolor{blue}{33.39}	& 30.84	& \textcolor{green}{29.42} & \textcolor{red}{30.51} & 29.10	& \textcolor{blue}{29.49}	& \textcolor{red}{32.72}	& \textcolor{green}{30.49}	& \textcolor{blue}{30.35}	& \textcolor{red}{30.14}	&  \textcolor{blue}{30.26}	& \textcolor{green}{30.57}\\
\cline{1-15}
\multirow{22}*{$\sigma$=50} & BM3D \cite{Dabov2007} & 26.13 & 29.69 & 26.68 & 25.04 & 25.82 & 25.10 & 25.90 & 29.05 & \textcolor{green}{27.22} & 26.78 & 26.81 & 26.46 & 26.72\\
\cline{2-15}
    & WNNM \cite{Gu2014} & 26.45 & 30.33 & 26.95 & 25.44 & 26.32 & 25.42 & 26.14 & 29.25 & \textcolor{red}{27.79} & 26.97 & 26.94 & 26.64 & 27.05\\
\cline{2-15}
    & TNRD \cite{Chen2017} & 	26.62 & 29.48 & 27.10 & 25.42 & 26.31 & 25.59 & 26.16 & 28.93 & 25.70 & 26.94 & 26.98 & 26.50 & 26.81\\
\cline{2-15}
    & DnCNN-S \cite{Zhang2017} & 27.03 & 30.00 & 27.32 & 25.70 & 26.78 & 25.87 & 26.48 & 29.39 & 26.22 & 27.20 & 27.24 & 26.90 & 27.18\\
\cline{2-15}
    & IRCNN \cite{ZhangZGZ2017} & 26.88 & 29.96 & 27.33 & 25.57 & 26.61 & \textcolor{green}{25.89} & 26.55 & 29.40 & 26.24 & 27.17 & 27.17 & 26.88 & 27.14 \\
\cline{2-15}
    & BUIFD \cite{Helou2020} & 25.44	& 29.76	& 26.50	& 24.87	& 26.49	& 25.34	& 25.07	& 28.81	& 25.49	& 26.59	& 26.87	& 26.34	& 26.46\\
\cline{2-15}
    & FFDNet \cite{Zhang2018} & 27.05 & 30.37 & 27.54 & 25.75 & 26.81 & \textcolor{green}{25.89} & \textcolor{green}{26.57} & \textcolor{blue}{29.66} & 26.45 & 27.33 & \textcolor{blue}{27.29} & \textcolor{green}{27.08} & 27.32 \\
\cline{2-15}
    & BRDNet \cite{Tian2020} & \textcolor{red}{27.44} & \textcolor{green}{30.53} & \textcolor{blue}{27.67} & \textcolor{green}{25.77} & \textcolor{blue}{26.97} & \textcolor{blue}{25.93} & \textcolor{blue}{26.66} & 25.93 & 26.66 & \textcolor{blue}{27.38} & \textcolor{green}{27.27} & \textcolor{blue}{27.17} & \textcolor{blue}{27.45}\\
\cline{2-15}
    & ADNet \cite{TianX2020} & \textcolor{green}{27.31} & \textcolor{blue}{30.59} & \textcolor{red}{27.69} & 25.70 & 26.90 & 25.88 & 26.56 & \textcolor{green}{29.59} & 26.64 & \textcolor{green}{27.35} & 27.17 & 27.07 & \textcolor{green}{27.37}\\
\cline{2-15}
    & DudeNet \cite{Tian2021} & 27.22 & 30.27 & 27.51 & \textcolor{blue}{25.88} & \textcolor{green}{26.93} & 25.88 & 26.50 & 29.45 & 26.49 & 27.26 & 27.19 & 26.97 & 27.30\\
\cline{2-15}
    & DCANet & \textcolor{blue}{27.41}	& \textcolor{red}{30.64}	& \textcolor{green}{27.57}	& \textcolor{red}{25.89}	& \textcolor{red}{27.08}	& \textcolor{red}{25.97}	& \textcolor{red}{26.67}	& \textcolor{red}{29.87}	& \textcolor{blue}{27.33}	& \textcolor{red}{27.47}	& \textcolor{red}{27.33}	& \textcolor{red}{27.25}	& \textcolor{red}{27.54}\\
\cline{1-15}
\end{tabular}
\end{table*}

Table \ref{tab:Set12_SSIM} presents the average SSIM results of the compared models on the Set12 dataset. As can be seen from the table, our DCANet also produces competitive results.

%SSIM results of Set12 dataset at different noise levels
\begin{table}[htbp]\scriptsize
\centering
\caption{Quantitative comparison results on the Set12 dataset. The three best results are respectively emphasized in red, blue and green.}
\label{tab:Set12_SSIM}
\begin{tabular}{cccc}
\cline{1-4}
Models & $\sigma$=15 & $\sigma$=25 & $\sigma$=50 \\
\cline{1-4}
BM3D \cite{Dabov2007} & 0.896 & 0.851 & 0.766\\
\cline{1-4}
WNNM \cite{Gu2014} & 0.894 & 0.846 & 0.756\\
\cline{1-4}
TNRD \cite{Chen2017} & 0.896 &  0.851 & 0.768 \\
\cline{1-4}
DnCNN-S \cite{Zhang2017} & \textcolor{green}{0.903} & 0.862 & 0.783 \\
\cline{1-4}
IRCNN  \cite{ZhangZGZ2017} & 0.901 & 0.860 & 0.780 \\
\cline{1-4}
BUIFD \cite{Helou2020} &  0.899 & 0.855 & 0.755 \\
\cline{1-4}
FFDNet \cite{Zhang2018} & \textcolor{green}{0.903} & 0.864 & 0.791 \\
\cline{1-4}
BRDNet \cite{Tian2020} & \textcolor{red}{0.906} & \textcolor{blue}{0.866} & \textcolor{blue}{0.794} \\
\cline{1-4}
ADNet \cite{TianX2020} & \textcolor{blue}{0.905} & \textcolor{green}{0.865} & 0.791 \\
\cline{1-4}
RIDNet \cite{Anwar2019} & \textcolor{red}{0.906} & \textcolor{red}{0.867} & \textcolor{green}{0.793} \\
\cline{1-4}
DCANet & \textcolor{green}{0.903} & \textcolor{blue}{0.866} & \textcolor{red}{0.798} \\
\cline{1-4}
\end{tabular}
\end{table}

Table \ref{tab:BSD68} reports the quantitative results on the BSD68 dataset of the compared denoising models. It can be found that the RIDNet obtains leading performance at noise levels 15 and 25. Our proposed model can obtain competitive denoising results at all tested noise levels, and especially our model outperforms other denoising methods at noise level 50.

%PSNR and SSIM results of BSD68 at different noise levels
\begin{table}[htbp]\scriptsize
\centering
\caption{Quantitative comparison results on the BSD68 dataset. The three best results are respectively emphasized in red, blue and green.}
\label{tab:BSD68}
\begin{tabular}{ccccc}
\cline{1-5}
Metrics & Models & $\sigma$=15 & $\sigma$=25 & $\sigma$=50\\
\cline{1-5}
\multirow{26}*{PSNR} & BM3D \cite{Dabov2007} & 31.07 & 28.57 & 25.62\\
\cline{2-5}
    & WNNM \cite{Gu2014} & 31.37 	& 28.83 & 25.87\\
\cline{2-5}
    & DnCNN-S \cite{Zhang2017} & 31.72 & 29.23 & 26.23\\
\cline{2-5}
    & TNRD \cite{Chen2017} & 31.42 & 28.92 & 25.97\\
\cline{2-5}
    & IRCNN \cite{ZhangZGZ2017} & 31.63 & 29.15 & 26.19 \\
\cline{2-5}
    & BUIFD \cite{Helou2020} &  31.35 & 28.75 & 25.11\\
\cline{2-5}
    & FFDNet \cite{Zhang2018} & 31.63	& 29.19 & 26.29\\
\cline{2-5}
    & DSNetB \cite{Peng2019} & 31.69 & 29.22 & 26.29 \\
\cline{2-5}
    & RIDNet \cite{Anwar2019} & \textcolor{red}{31.81} & \textcolor{red}{29.34} & \textcolor{blue}{26.40} \\
\cline{2-5}
    & BRDNet \cite{Tian2020} & \textcolor{blue}{31.79} & \textcolor{green}{29.29} & \textcolor{green}{26.36} \\
\cline{2-5}
    & ADNet \cite{TianX2020} & 31.74 & 29.25 & 26.29 \\
\cline{2-5}
    & AINDNet \cite{Kim2020} & 31.69 & 29.26 & 26.32 \\
\cline{2-5}
    & DudeNet \cite{Tian2021} & \textcolor{green}{31.78} & \textcolor{green}{29.29} & 26.31 \\
\cline{2-5}
    & DCANet & 31.71 & \textcolor{blue}{29.31} & \textcolor{red}{26.44}\\
\cline{1-5}
\multirow{20}*{SSIM} & BM3D \cite{Dabov2007} & 0.872 & 0.802 & 0.687\\
\cline{2-5}
    & WNNM \cite{Gu2014} & 0.878 & 0.810 & 0.698\\
\cline{2-5}
    & TNRD \cite{Chen2017} & 0.883 & 0.816 & 0.703\\
\cline{2-5}
    & DnCNN-S \cite{Zhang2017} & \textcolor{green}{0.891} & 0.828 & 0.719 \\
\cline{2-5}
    & IRCNN  \cite{ZhangZGZ2017} & 0.888 & 0.825 & 0.717 \\
\cline{2-5}
    & BUIFD \cite{Helou2020} &  0.886 & 0.819 & 0.682 \\
\cline{2-5}
    & FFDNet \cite{Zhang2018} & 0.890 & \textcolor{green}{0.830} & \textcolor{green}{0.726} \\
\cline{2-5}
    & BRDNet \cite{Tian2020} & \textcolor{red}{0.893} & \textcolor{blue}{0.831} & \textcolor{blue}{0.727} \\
\cline{2-5}
    & ADNet  \cite{TianX2020} & \textcolor{blue}{0.892} & 0.829 &  0.722 \\
\cline{2-5}
    & RIDNet \cite{Anwar2019} &  \textcolor{red}{0.893} & \textcolor{red}{0.833} & \textcolor{blue}{0.727} \\
\cline{2-5}
    & DCANet & \textcolor{green}{0.891} & \textcolor{blue}{0.831} & \textcolor{red}{0.730} \\
\cline{1-5}
\end{tabular}
\end{table}

\begin{figure*}[htbp]
	\centering
    \captionsetup[subfigure]{labelformat=empty}
	\begin{subfigure}{0.18\linewidth}
		\centering
		\includegraphics[width=0.99\linewidth]{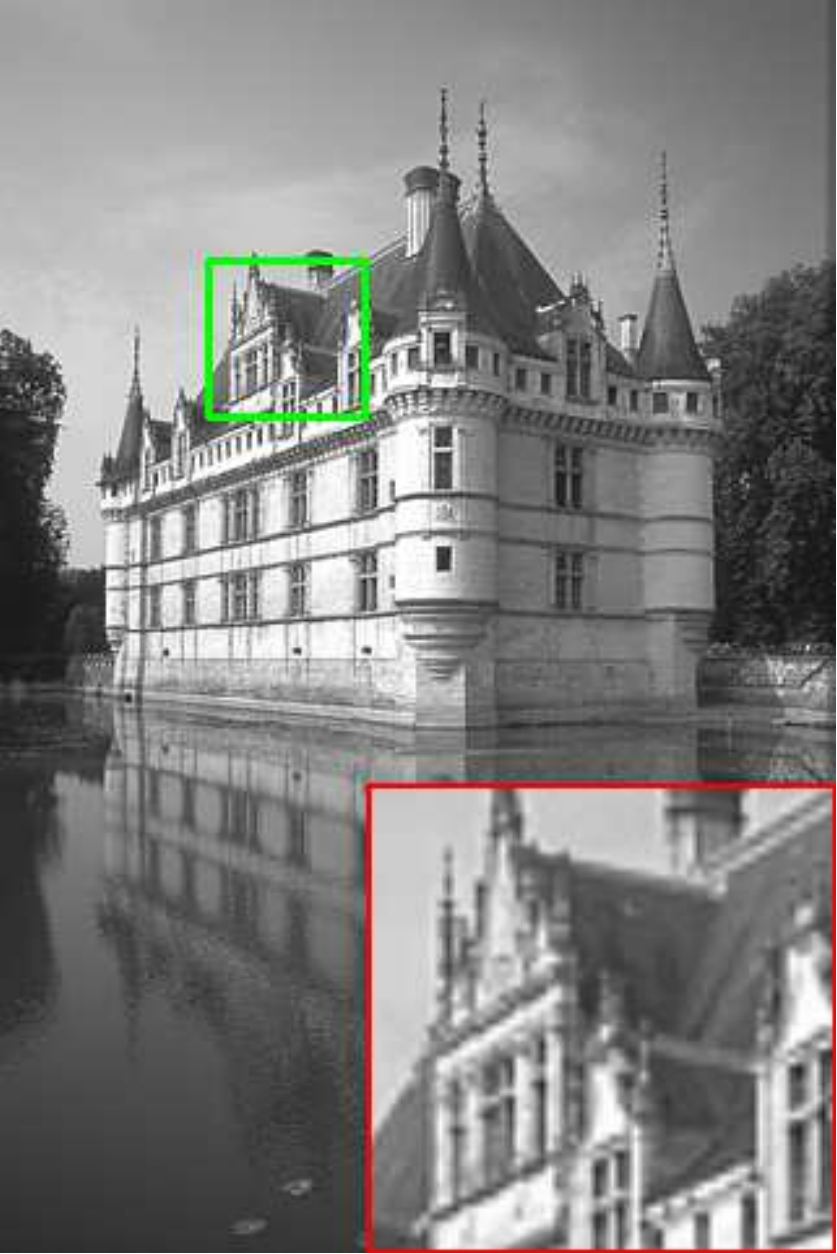}
		\caption{(\uppercase\expandafter{\romannumeral1})}
	\end{subfigure}
    \centering
	\begin{subfigure}{0.18\linewidth}
		\centering
		\includegraphics[width=0.99\linewidth]{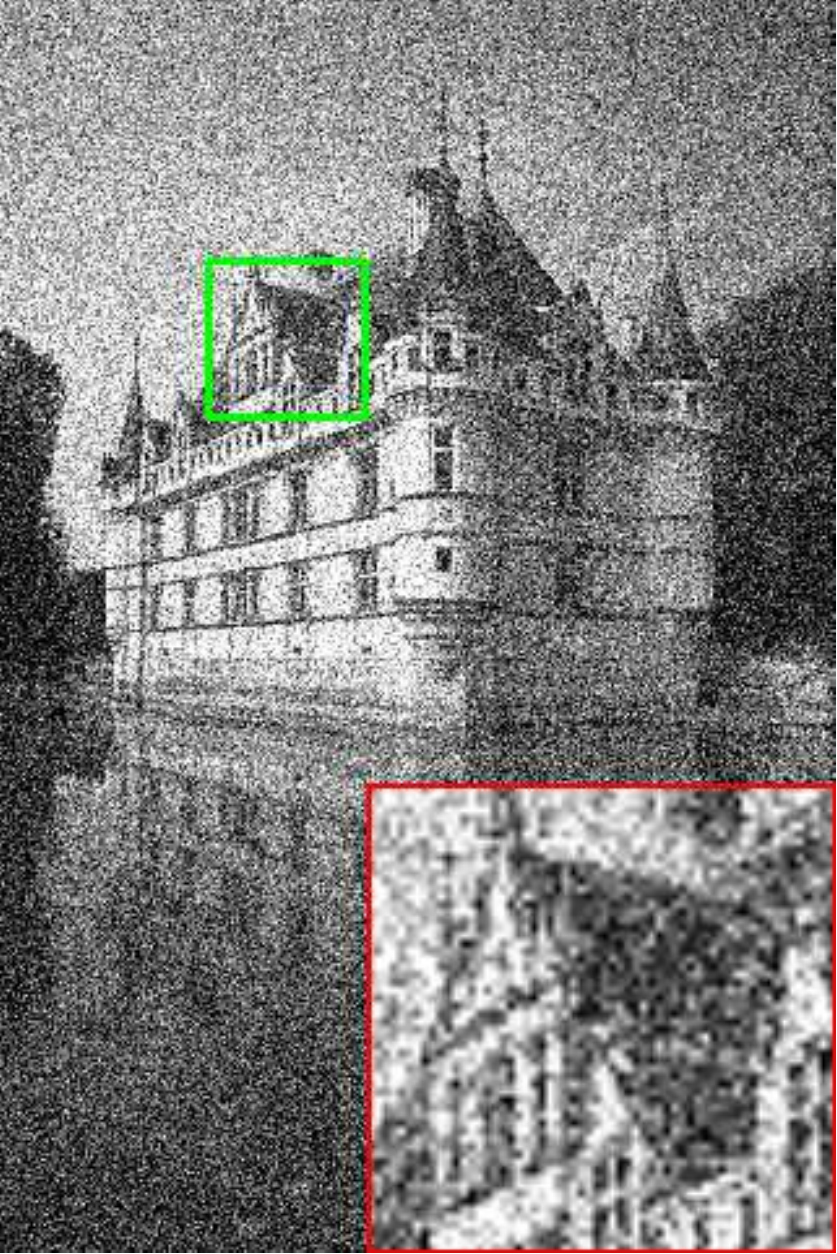}
		\caption{(\uppercase\expandafter{\romannumeral2})}
	\end{subfigure}
    \centering
	\begin{subfigure}{0.18\linewidth}
		\centering
		\includegraphics[width=0.99\linewidth]{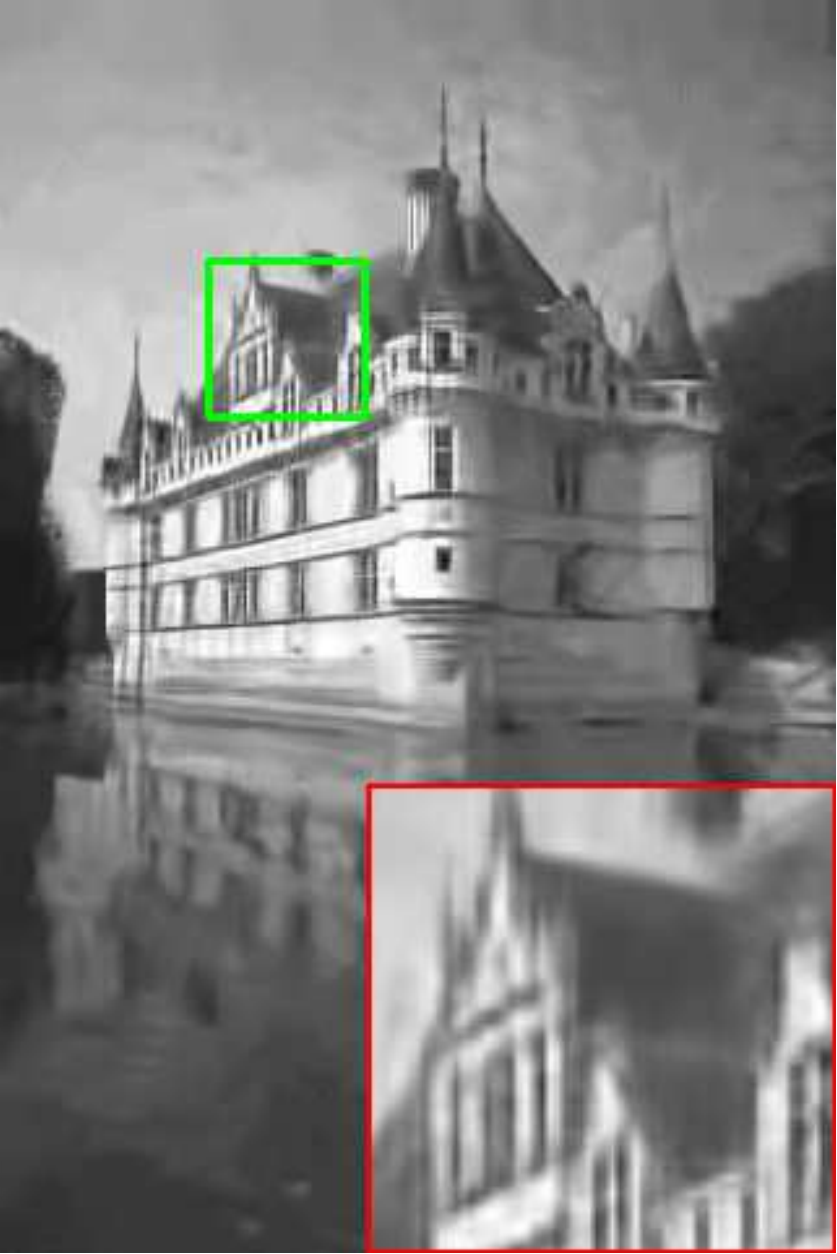}
		\caption{(\uppercase\expandafter{\romannumeral3})}
	\end{subfigure}
    \centering
	\begin{subfigure}{0.18\linewidth}
		\centering
		\includegraphics[width=0.99\linewidth]{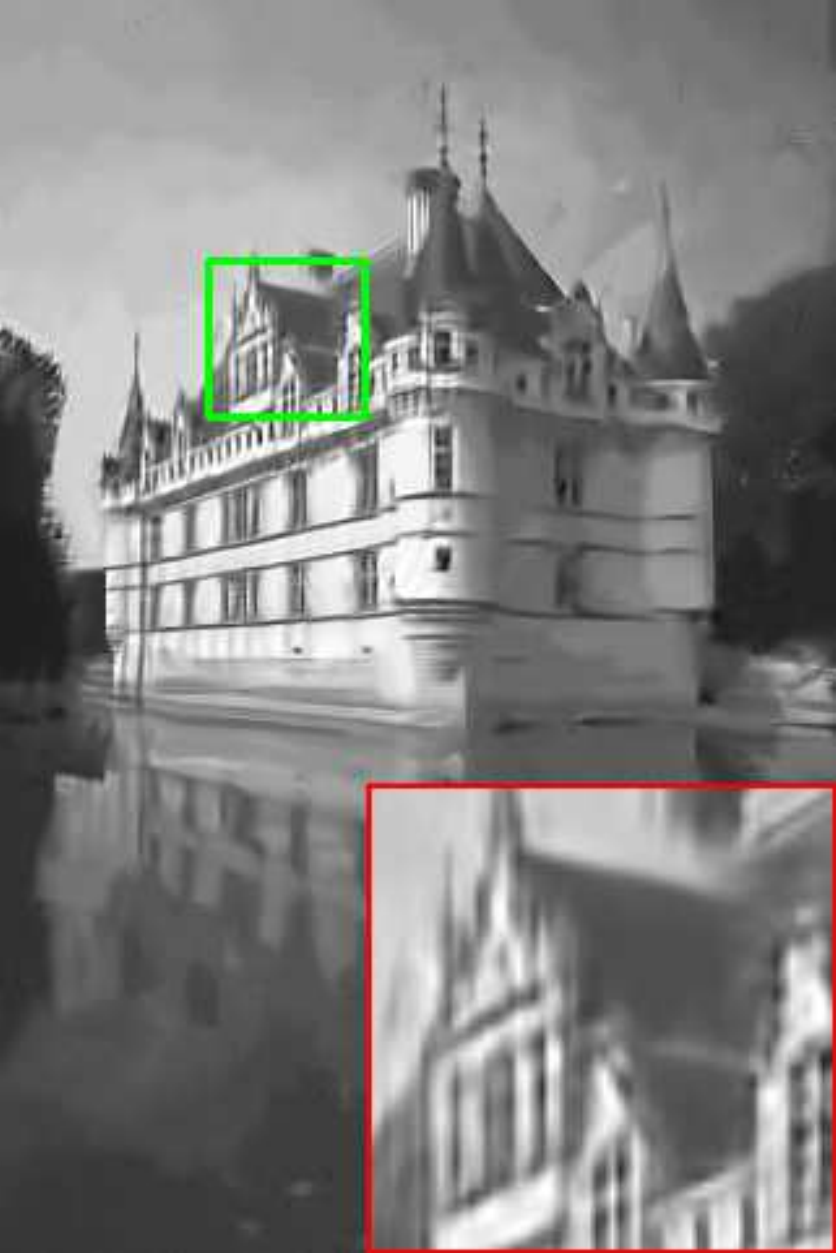}
		\caption{(\uppercase\expandafter{\romannumeral4})}
	\end{subfigure}
    \centering
	\begin{subfigure}{0.18\linewidth}
		\centering
		\includegraphics[width=0.99\linewidth]{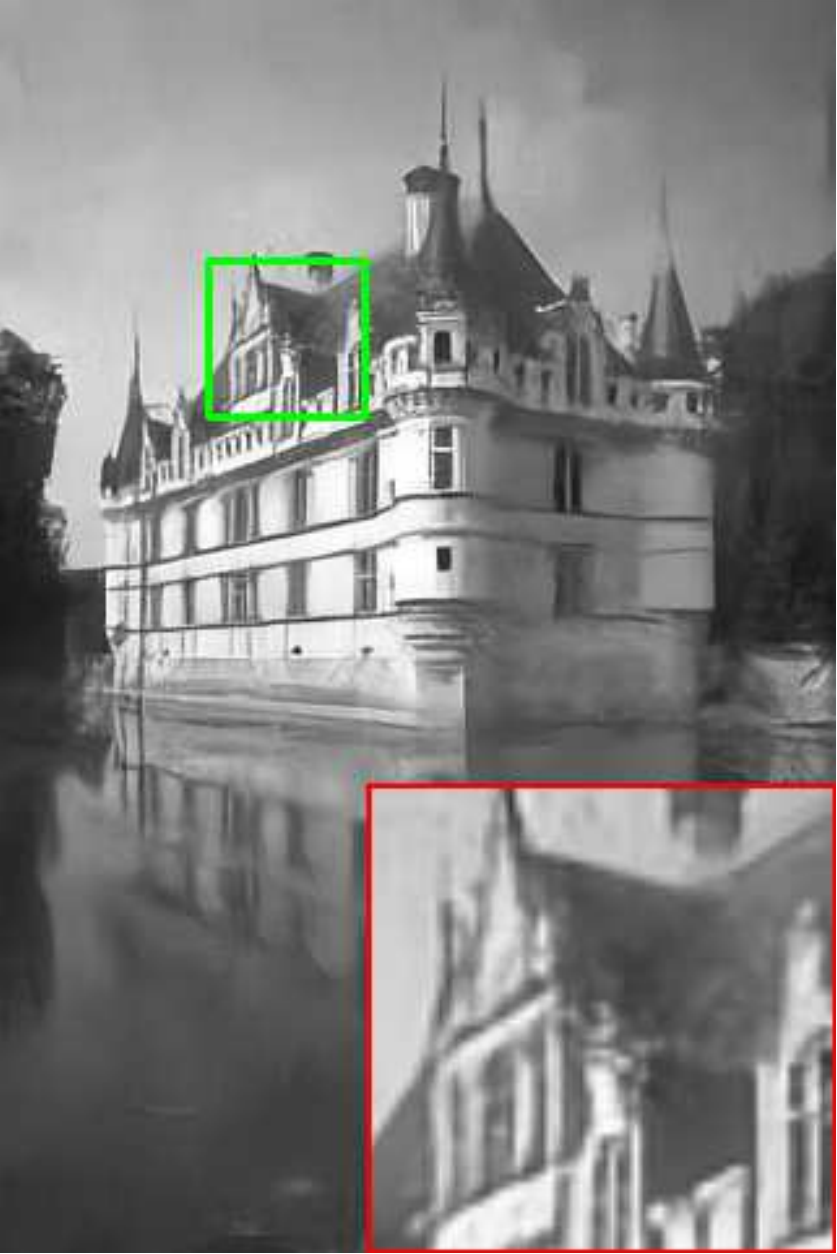}
		\caption{(\uppercase\expandafter{\romannumeral5})}
	\end{subfigure}
    \centering
	\begin{subfigure}{0.18\linewidth}
		\centering
		\includegraphics[width=0.99\linewidth]{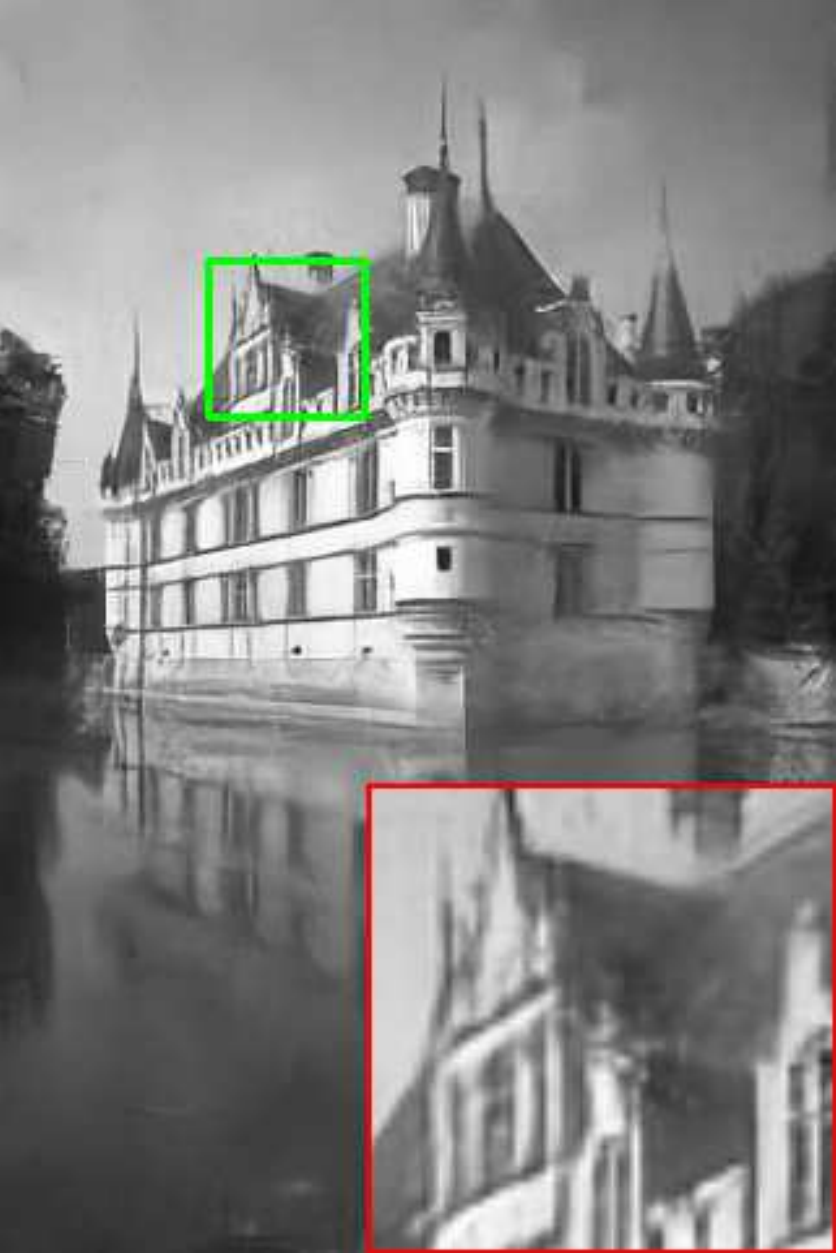}
		\caption{(\uppercase\expandafter{\romannumeral6})}
	\end{subfigure}
    \centering
	\begin{subfigure}{0.18\linewidth}
		\centering
		\includegraphics[width=0.99\linewidth]{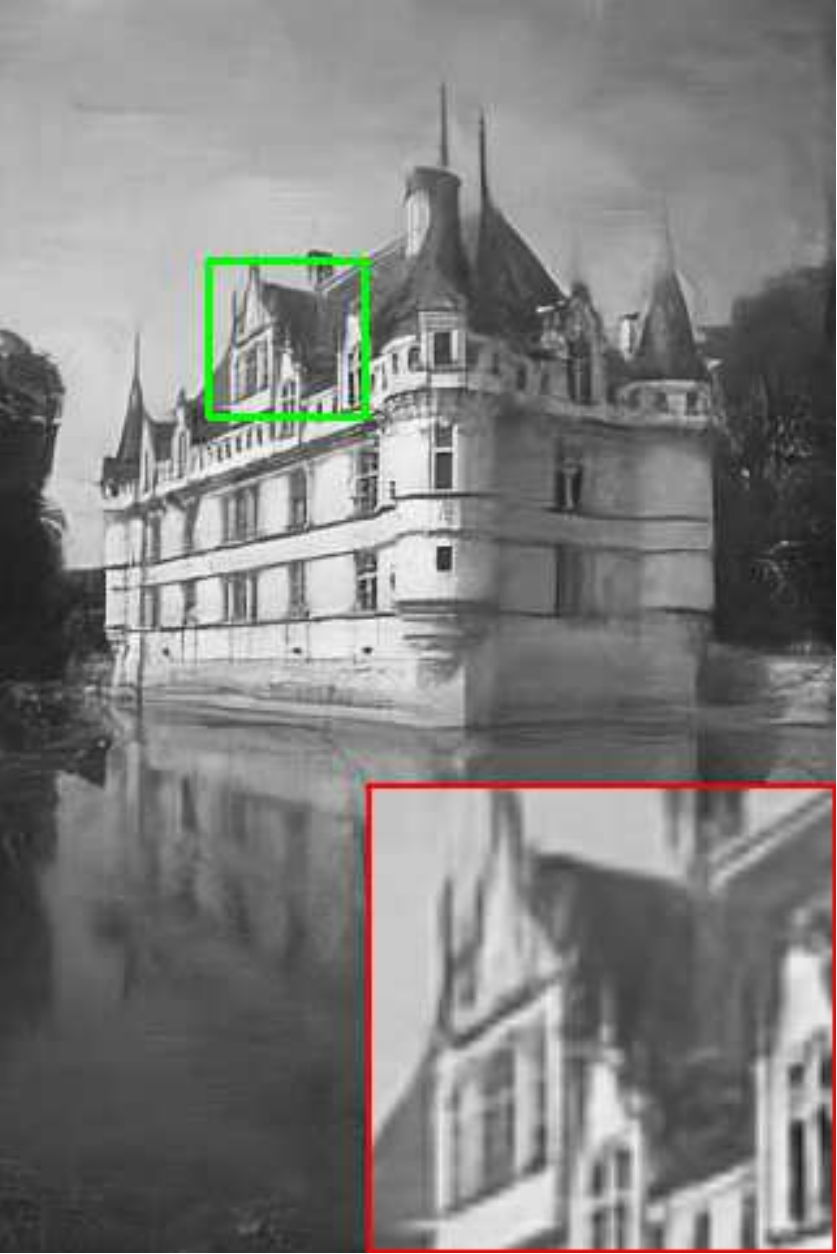}
		\caption{(\uppercase\expandafter{\romannumeral7})}
	\end{subfigure}
    \centering
	\begin{subfigure}{0.18\linewidth}
		\centering
		\includegraphics[width=0.99\linewidth]{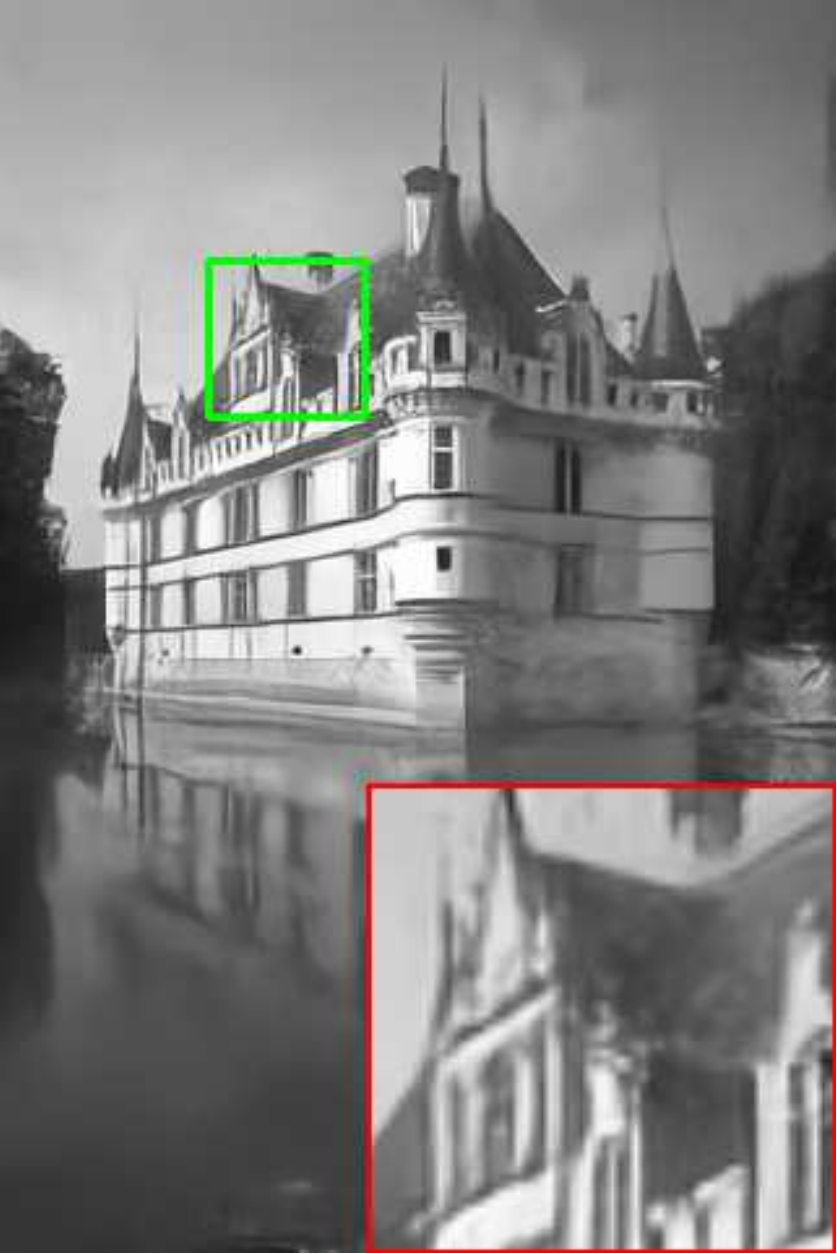}
		\caption{(\uppercase\expandafter{\romannumeral8})}
	\end{subfigure}
    \centering
	\begin{subfigure}{0.18\linewidth}
		\centering
		\includegraphics[width=0.99\linewidth]{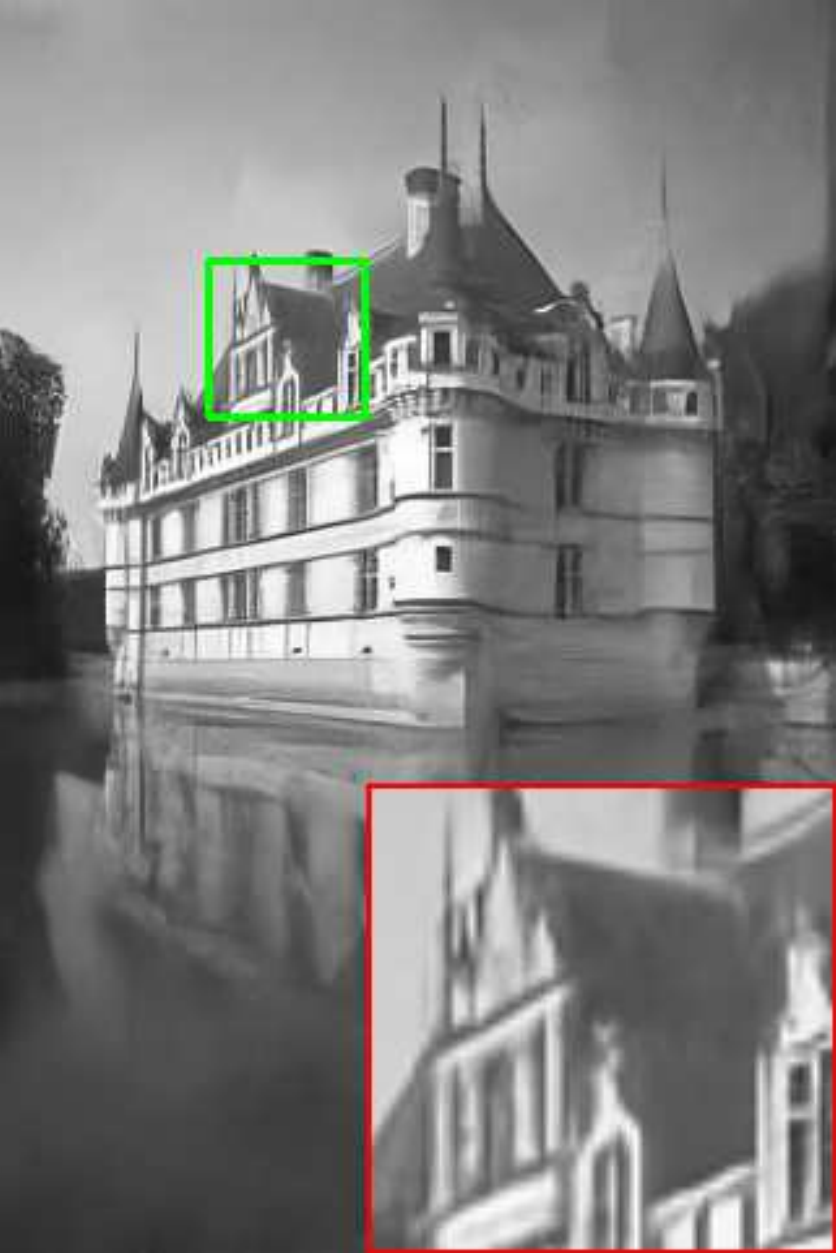}
		\caption{(\uppercase\expandafter{\romannumeral9})}
	\end{subfigure}
    \centering
	\begin{subfigure}{0.18\linewidth}
		\centering
		\includegraphics[width=0.99\linewidth]{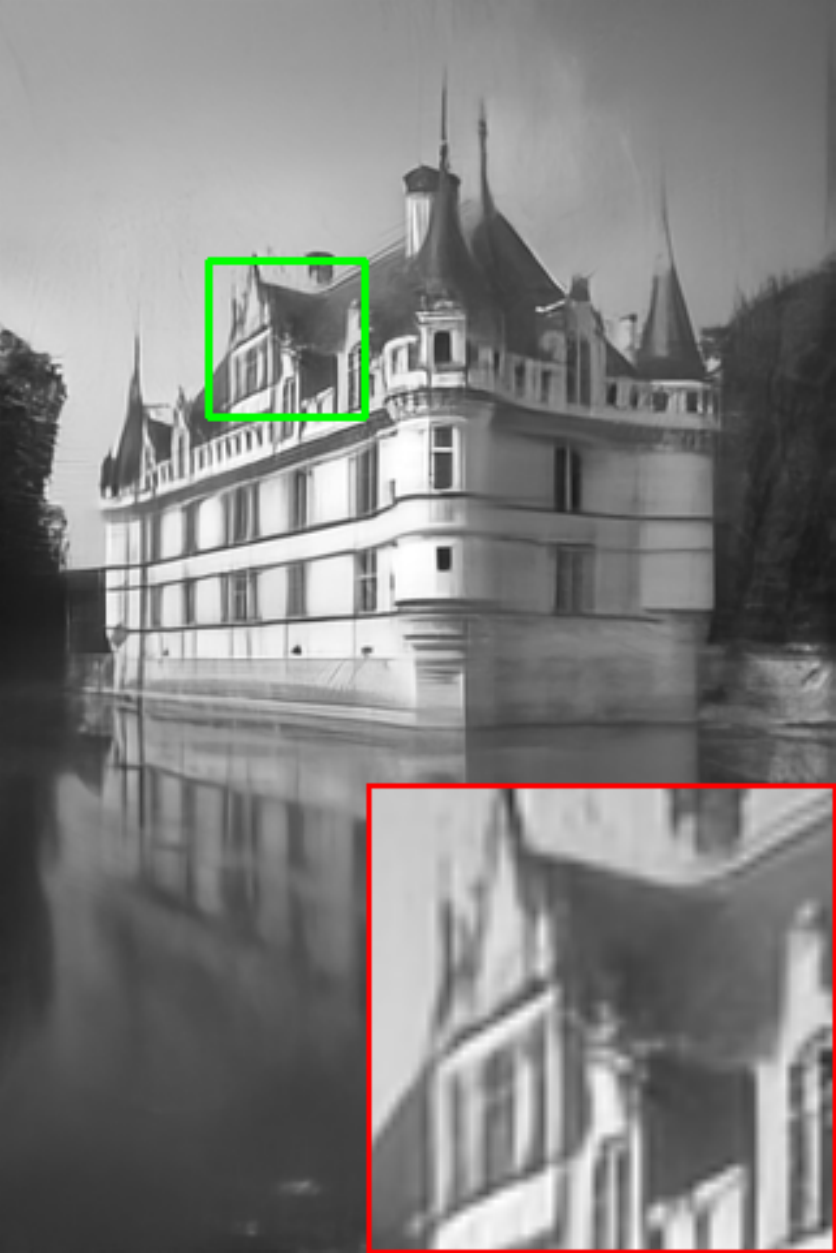}
		\caption{(\uppercase\expandafter{\romannumeral10})}
	\end{subfigure}
\caption{Visual comparison results on the ``test003'' image from the BSD68 dataset at the noise level 50. (\uppercase\expandafter{\romannumeral1}) Ground-truth image, (\uppercase\expandafter{\romannumeral2}) Noisy image / 14.15dB, (\uppercase\expandafter{\romannumeral3}) BM3D / 26.21dB, (\uppercase\expandafter{\romannumeral4}) WNNM / 26.50dB, (\uppercase\expandafter{\romannumeral5}) DnCNN-S / 26.90dB, (\uppercase\expandafter{\romannumeral6}) IRCNN / 26.85dB, (\uppercase\expandafter{\romannumeral7}) BUIFD / 26.32dB, (\uppercase\expandafter{\romannumeral8}) FFDNet / 26.92dB, (\uppercase\expandafter{\romannumeral9}) ADNet / 27.06dB, (\uppercase\expandafter{\romannumeral10}) DCANet / 27.18dB.}
\label{fig:Castle}
\end{figure*}

Fig. \ref{fig:Castle} shows the visual comparison results on the ``test003'' image at noise level 50 of different denoising models. One can find that compared with other models, the DCANet obtains better denoising performance, both quantitatively and qualitatively.

It can be seen from our experimental results on the grayscale images that our DCANet can achieve competitive denoising performance compared with other state-of-the-art methods. Especially, a trend can be discovered from Table \ref{tab:Set12_PSNR}, Table \ref{tab:Set12_SSIM}, and Table \ref{tab:BSD68}, that for grayscale image denoising, our model becomes more effective as the noise power increases.

%Denoising results on the image ``test003'' from Set12 dataset with noise level of 50.

For the noise removal evaluation on color images, the CBSD68, Kodak24, and McMaster datasets were applied. Table \ref{tab:color_PSNR} lists the average PSNR values at different noise levels of many models on three datasets. It can be seen that our model obtains leading denoising results compared to other state-of-the-art denoising models.

%The average PSNR (dB) on CBSD68, Kodak24 and McMaster datasets.
\begin{table*}[htbp]\scriptsize
\centering
\caption{Quantitative comparison results on three commonly used color datasets. The three best results are respectively emphasized in red, blue and green.}
\label{tab:color_PSNR}
\begin{tabular}{ccccccc}
\cline{1-7}
Datasets &  Models & $\sigma$=15 & $\sigma$=25 & $\sigma$=35 & $\sigma$=50 & $\sigma$=75\\
\cline{1-7}
\multirow{24}*{CBSD68} & CBM3D \cite{Dabov2007} & 33.52 & 30.71 & 28.89 & 27.38 & 25.74\\
\cline{2-7}
    & MCWNNM \cite{Xu2017} &  30.91 & 27.61 & 25.88 & 23.18 & 21.21\\
\cline{2-7}
    & CDnCNN-S \cite{Zhang2017} & 33.89 & 31.23 & 29.58 & 27.92 & 24.47\\
\cline{2-7}
    & IRCNN \cite{ZhangZGZ2017} & 33.86 & 31.16 & 29.50 & 27.86 & -\\
\cline{2-7}
    & BUIFD \cite{Helou2020} & 33.65 & 30.76 & 28.82 & 26.61 & 23.64\\
\cline{2-7}
    & FFDNet \cite{Zhang2018} & 33.87 & 31.21 & 29.58 & 27.96 & 26.24\\
\cline{2-7}
    & DSNetB \cite{Peng2019} & 33.91 & 31.28 & - & 28.05 & - \\
\cline{2-7}
    & RIDNet \cite{Anwar2019} & \textcolor{green}{34.01} & \textcolor{green}{31.37} & - & \textcolor{green}{28.14} & -\\
\cline{2-7}
    & BRDNet \cite{Tian2020} & \textcolor{red}{34.10} & \textcolor{blue}{31.43} & \textcolor{blue}{29.77} & \textcolor{blue}{28.16} & \textcolor{blue}{26.43}\\
\cline{2-7}
    & ADNet \cite{TianX2020} & 33.99 & 31.31 & 29.66 & 28.04 & 26.33\\
\cline{2-7}
    & DudeNet \cite{Tian2021} & \textcolor{green}{34.01} & 31.34 & \textcolor{green}{29.71} & 28.09 & \textcolor{green}{26.40}\\
\cline{2-7}
    & AirNet \cite{Li2022} & 33.92 & 31.26 & - & 28.01 & -\\
\cline{2-7}
    & DCANet & \textcolor{blue}{34.05} & \textcolor{red}{31.45} & \textcolor{red}{29.86} & \textcolor{red}{28.28} & \textcolor{red}{26.61}\\
\cline{1-7}
\multirow{22}*{Kodak24} & CBM3D \cite{Dabov2007} & 34.28 & 31.68 & 29.90 & 28.46 & 26.82\\
\cline{2-7}
    & MCWNNM \cite{Xu2017} & 32.00 & 28.76 & 27.02 & 21.18 & 18.06\\
\cline{2-7}
    & CDnCNN-S \cite{Zhang2017} & 34.48 & 32.03 & 30.46 & 28.85 & 25.04\\
\cline{2-7}
    & IRCNN \cite{ZhangZGZ2017} & 34.56 & 32.03 & 30.43 & 28.81 & -\\
\cline{2-7}
    & BUIFD \cite{Helou2020} & 34.41 & 31.77 & 29.94 & 27.74 & 24.67\\
\cline{2-7}
    & FFDNet \cite{Zhang2018} & 34.63 & 32.13 & 30.57 & 28.98 & 27.27\\
\cline{2-7}
    & DSNetB \cite{Peng2019} & 34.63 & 32.16 & - & 29.05 & -\\
\cline{2-7}
    & BRDNet \cite{Tian2020} & \textcolor{red}{34.88} & \textcolor{blue}{32.41} & \textcolor{blue}{30.80} & \textcolor{blue}{29.22} & \textcolor{blue}{27.49}\\
\cline{2-7}
    & ADNet \cite{TianX2020} & 34.76 & \textcolor{green}{32.26} & 30.68 & \textcolor{green}{29.10} & \textcolor{green}{27.40}\\
\cline{2-7}
    & DudeNet \cite{Tian2021} & \textcolor{green}{34.81} & \textcolor{green}{32.26} & \textcolor{green}{30.69} & \textcolor{green}{29.10} & 27.39\\
\cline{2-7}
    & AirNet \cite{Li2022} & 34.68 & 32.21 & - & 29.06 & - \\
\cline{2-7}
    & DCANet & \textcolor{blue}{34.85} & \textcolor{red}{32.43} & \textcolor{red}{30.91} & \textcolor{red}{29.36} & \textcolor{red}{27.69} \\
\cline{1-7}
\multirow{20}*{McMaster} & CBM3D \cite{Dabov2007} & 34.06 & 31.66 & 29.92 & 28.51 & 26.79\\
\cline{2-7}
    & MCWNNM \cite{Xu2017} & 32.75 & 29.39 & 27.44 & 21.37 & 18.16\\
\cline{2-7}
    & CDnCNN-S \cite{Zhang2017} & 33.44 & 31.51 & 30.14 & 28.61 & 25.10\\
\cline{2-7}
    & IRCNN \cite{ZhangZGZ2017} & 34.58 & 32.18 & 30.59 & 28.91 & -	\\
\cline{2-7}
    & BUIFD \cite{Helou2020} & 33.84 & 31.06 & 28.87 & 26.20 & 22.75\\
\cline{2-7}
    & FFDNet \cite{Zhang2018} & 34.66 & 32.35 & 30.81 & 29.18 & 27.33\\
\cline{2-7}
    & DSNetB \cite{Peng2019} & 34.67 & 32.40 & - & 29.28 & -\\
\cline{2-7}
    & BRDNet \cite{Tian2020} & \textcolor{red}{35.08} & \textcolor{red}{32.75} & \textcolor{blue}{31.15} & \textcolor{blue}{29.52} & \textcolor{blue}{27.72}\\
\cline{2-7}
    & ADNet \cite{TianX2020} & \textcolor{blue}{34.93} & 32.56 & \textcolor{green}{31.00} & \textcolor{green}{29.36} & \textcolor{green}{27.53}\\
\cline{2-7}
    & AirNet \cite{Li2022} & 34.70 & \textcolor{green}{32.44} & - & 29.26 & - \\
\cline{2-7}
    & DCANet & \textcolor{green}{34.84} & \textcolor{blue}{32.62} & \textcolor{red}{31.16} & \textcolor{red}{29.59} & \textcolor{red}{27.81}\\
\cline{1-7}
\end{tabular}
\end{table*}

Table \ref{tab:color_SSIM} reports the average SSIM values at multiple noise levels of the compared denoising methods on three color image datasets. One can find that the DCANet also generates better results.

%the SSIM results of CBSD68, Kodak24 and McMaster datasets at different noise levels.
\begin{table}[htbp]\scriptsize
\centering
\caption{Quantitative comparison results on three commonly used color datasets. The three best results are respectively emphasized in red, blue and green.}
\label{tab:color_SSIM}
\begin{tabular}{ccccc}
\cline{1-5}
Datasets & Models & $\sigma$=15 & $\sigma$=25 & $\sigma$=50\\
\cline{1-5}
\multirow{12}*{CBSD68} & CDnCNN-B \cite{Zhang2017} & 0.929 & \textcolor{green}{0.883} & 0.790\\
\cline{2-5}
    & IRCNN  \cite{ZhangZGZ2017} & 0.929 & 0.882 & 0.790\\
\cline{2-5}
    & BUIFD \cite{Helou2020} & \textcolor{green}{0.930} & 0.882 & 0.777\\
\cline{2-5}
    & FFDNet \cite{Zhang2018} & 0.929 & 0.882 & 0.789 \\
\cline{2-5}
    & ADNet \cite{TianX2020}  & \textcolor{red}{0.933} & \textcolor{red}{0.889} & \textcolor{green}{0.797}\\
\cline{2-5}
    & AirNet \cite{Li2022} & \textcolor{red}{0.933} & 0.\textcolor{blue}{888} & \textcolor{blue}{0.798}\\
\cline{2-5}
    & DCANet & \textcolor{blue}{0.932} & \textcolor{blue}{0.888} & \textcolor{red}{0.804}\\
\cline{1-5}
\multirow{12}*{Kodak24} & CDnCNN-B \cite{Zhang2017} & 0.920 & 0.876 & 0.791\\
\cline{2-5}
    & IRCNN  \cite{ZhangZGZ2017} & 0.920 & 0.877 & 0.793\\
\cline{2-5}
    & BUIFD \cite{Helou2020} & \textcolor{green}{0.923} & \textcolor{green}{0.879} & 0.786\\
\cline{2-5}
    & FFDNet \cite{Zhang2018} & 0.922 & 0.878 & 0.794\\
\cline{2-5}
    & ADNet \cite{TianX2020}  & \textcolor{blue}{0.924} & \textcolor{blue}{0.882} & \textcolor{green}{0.798}\\
\cline{2-5}
    & AirNet \cite{Li2022} & \textcolor{blue}{0.924} & \textcolor{blue}{0.882} & \textcolor{blue}{0.799}\\
\cline{2-5}
    & DCANet & \textcolor{red}{0.925} & \textcolor{red}{0.885} & \textcolor{red}{0.811}\\
\cline{1-5}
\multirow{12}*{McMaster} & CDnCNN-B \cite{Zhang2017} & 0.904 & 0.869 & 0.799\\
\cline{2-5}
    & IRCNN  \cite{ZhangZGZ2017} & 0.920 & 0.882 & 0.807\\
\cline{2-5}
    & BUIFD \cite{Helou2020} & 0.901 & 0.847 & 0.733\\
\cline{2-5}
    & FFDNet \cite{Zhang2018} & 0.922 & 0.886 & 0.815 \\
\cline{2-5}
    & ADNet \cite{TianX2020}  & \textcolor{red}{0.927} & \textcolor{red}{0.894} & \textcolor{blue}{0.825}\\
\cline{2-5}
    & AirNet \cite{Li2022} & \textcolor{blue}{0.925} & \textcolor{green}{0.891} & \textcolor{green}{0.822}\\
\cline{2-5}
    & DCANet & \textcolor{green}{0.924} & \textcolor{blue}{0.893} & \textcolor{red}{0.831}\\
\cline{1-5}
\end{tabular}
\end{table}

\begin{figure*}[htbp]
	\centering
    \captionsetup[subfigure]{labelformat=empty}
	\begin{subfigure}{0.15\linewidth}
		\centering
		\includegraphics[width=0.99\linewidth]{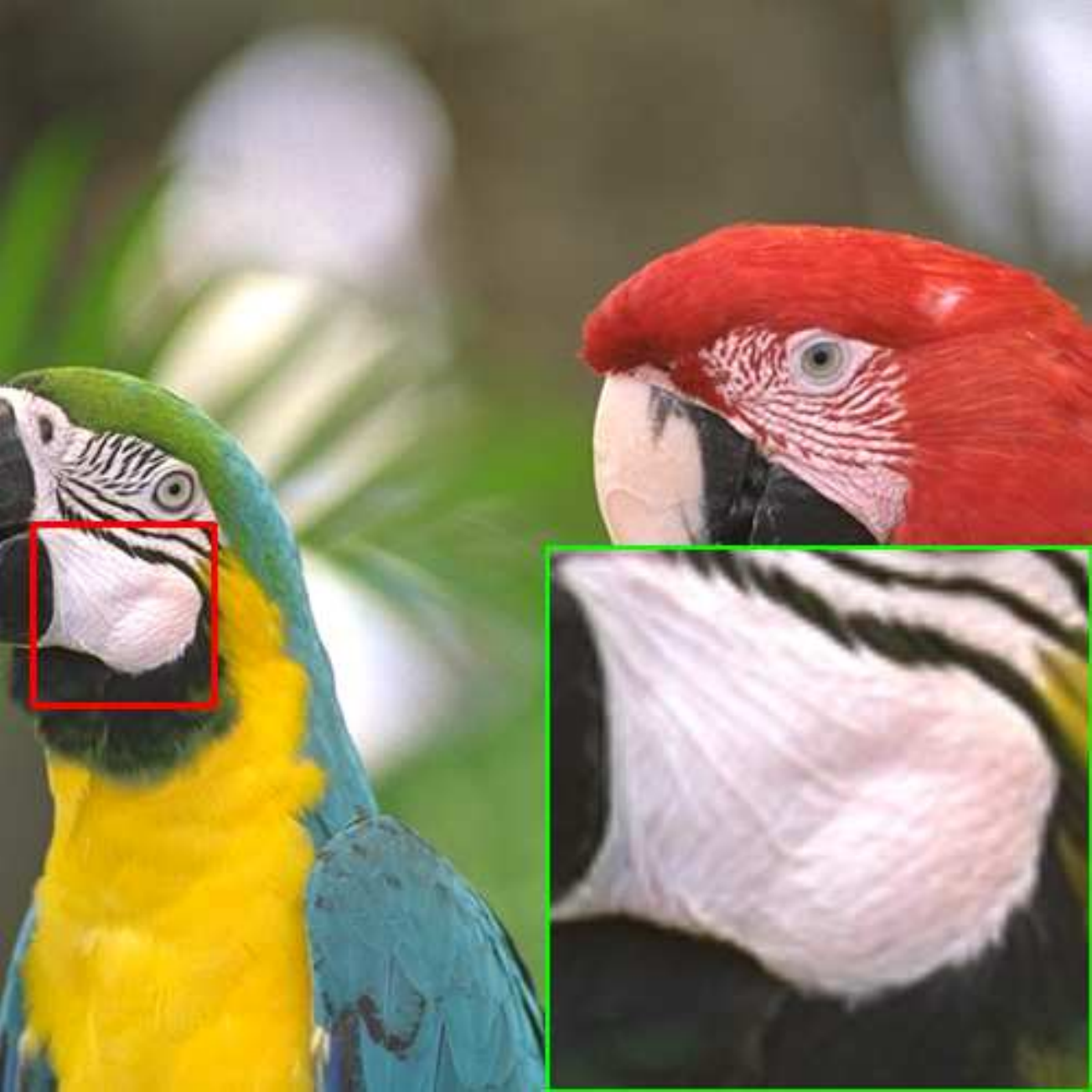}
		\caption{(\uppercase\expandafter{\romannumeral1})}
	\end{subfigure}
    \centering
	\begin{subfigure}{0.15\linewidth}
		\centering
		\includegraphics[width=0.99\linewidth]{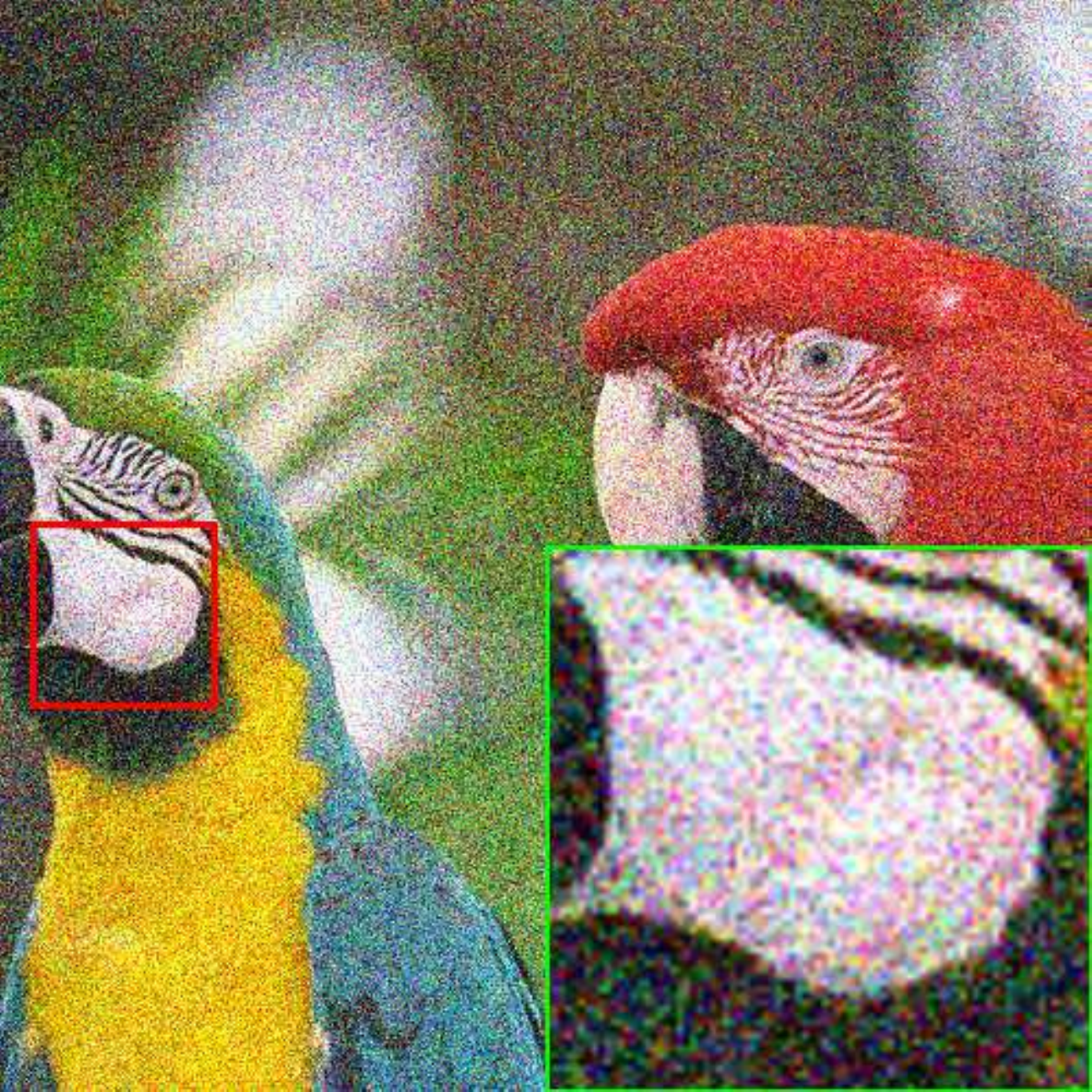}
		\caption{(\uppercase\expandafter{\romannumeral2})}
	\end{subfigure}
    \centering
	\begin{subfigure}{0.15\linewidth}
		\centering
		\includegraphics[width=0.99\linewidth]{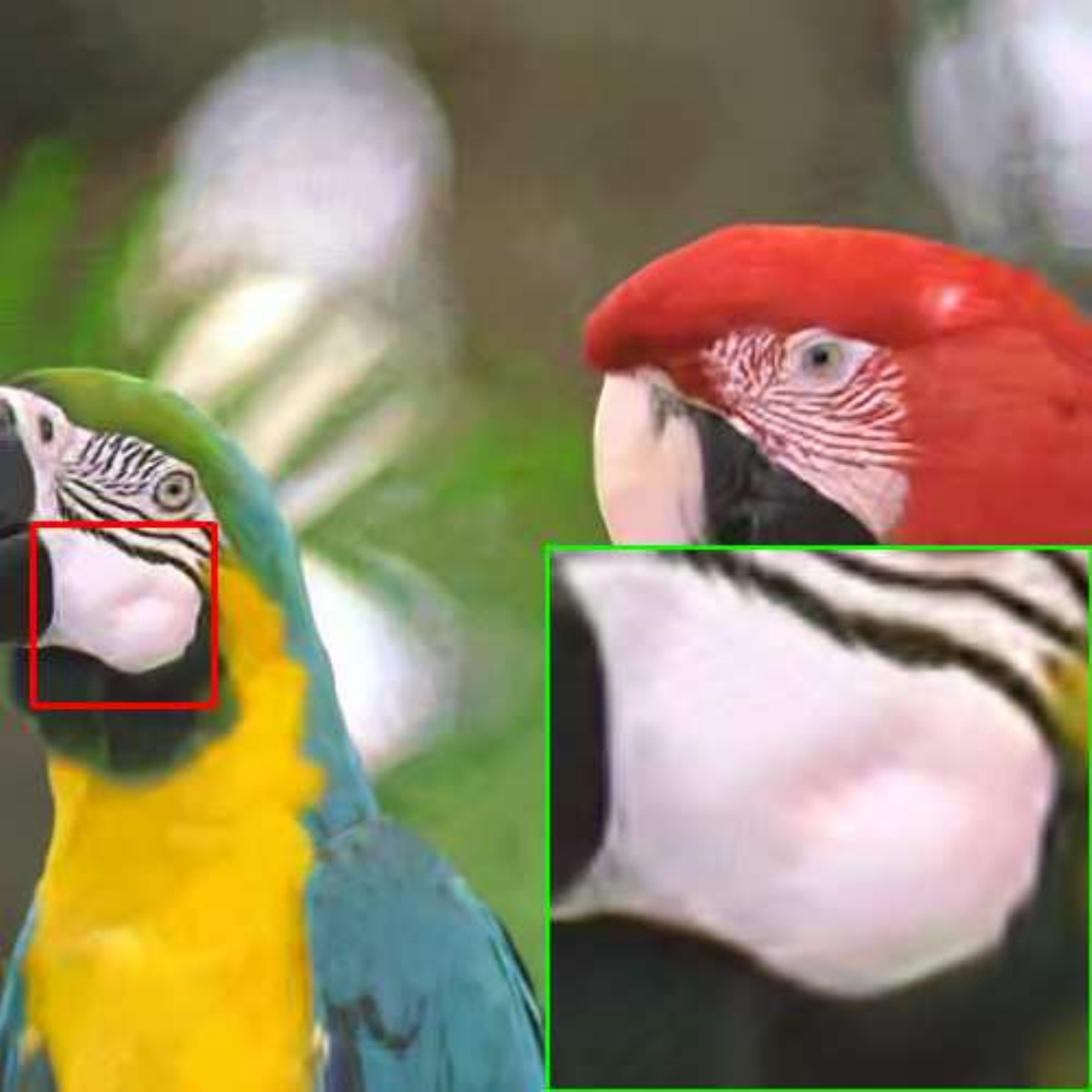}
		\caption{(\uppercase\expandafter{\romannumeral3})}
	\end{subfigure}
    \centering
	\begin{subfigure}{0.15\linewidth}
		\centering
		\includegraphics[width=0.99\linewidth]{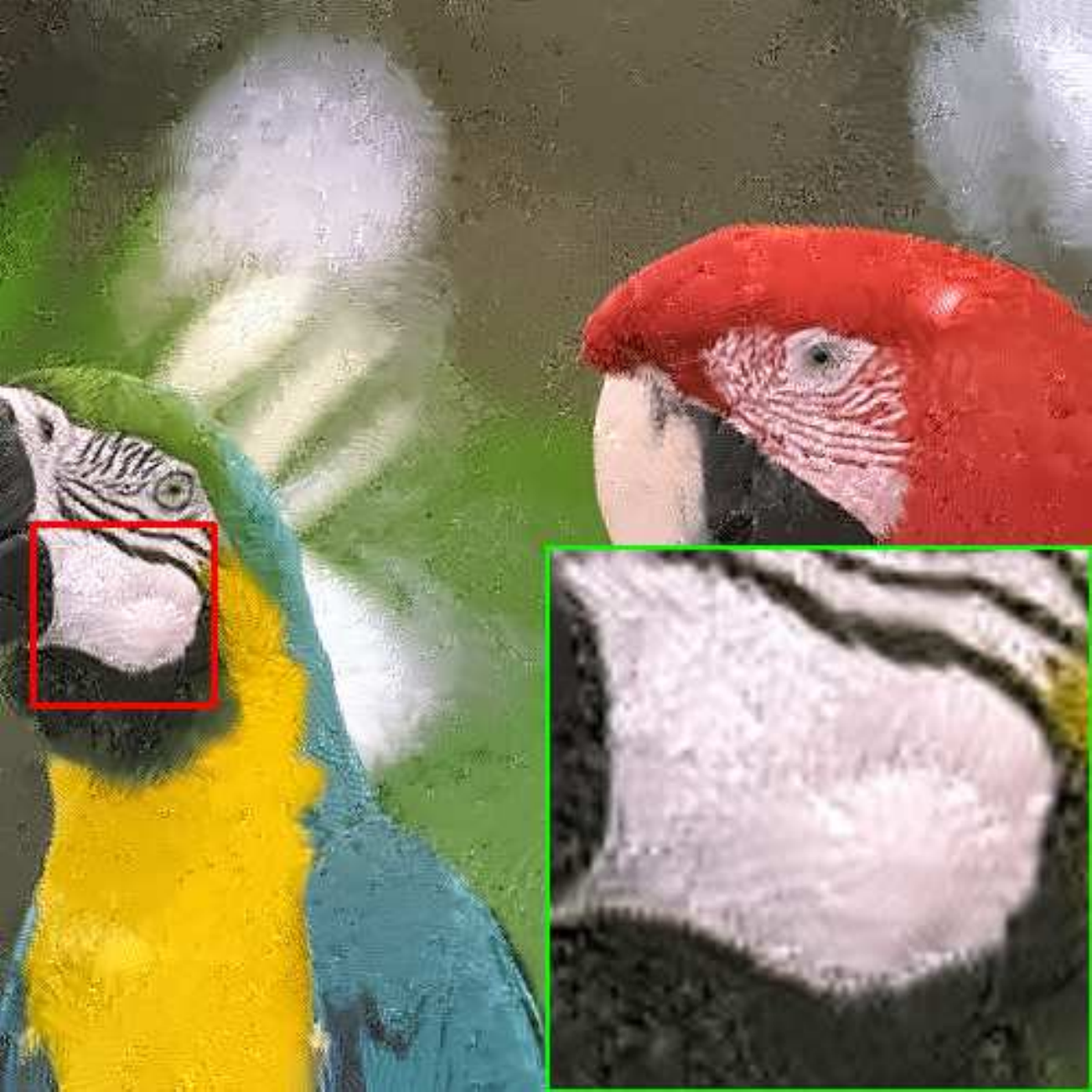}
		\caption{(\uppercase\expandafter{\romannumeral4})}
	\end{subfigure}
    \centering
	\begin{subfigure}{0.15\linewidth}
		\centering
		\includegraphics[width=0.99\linewidth]{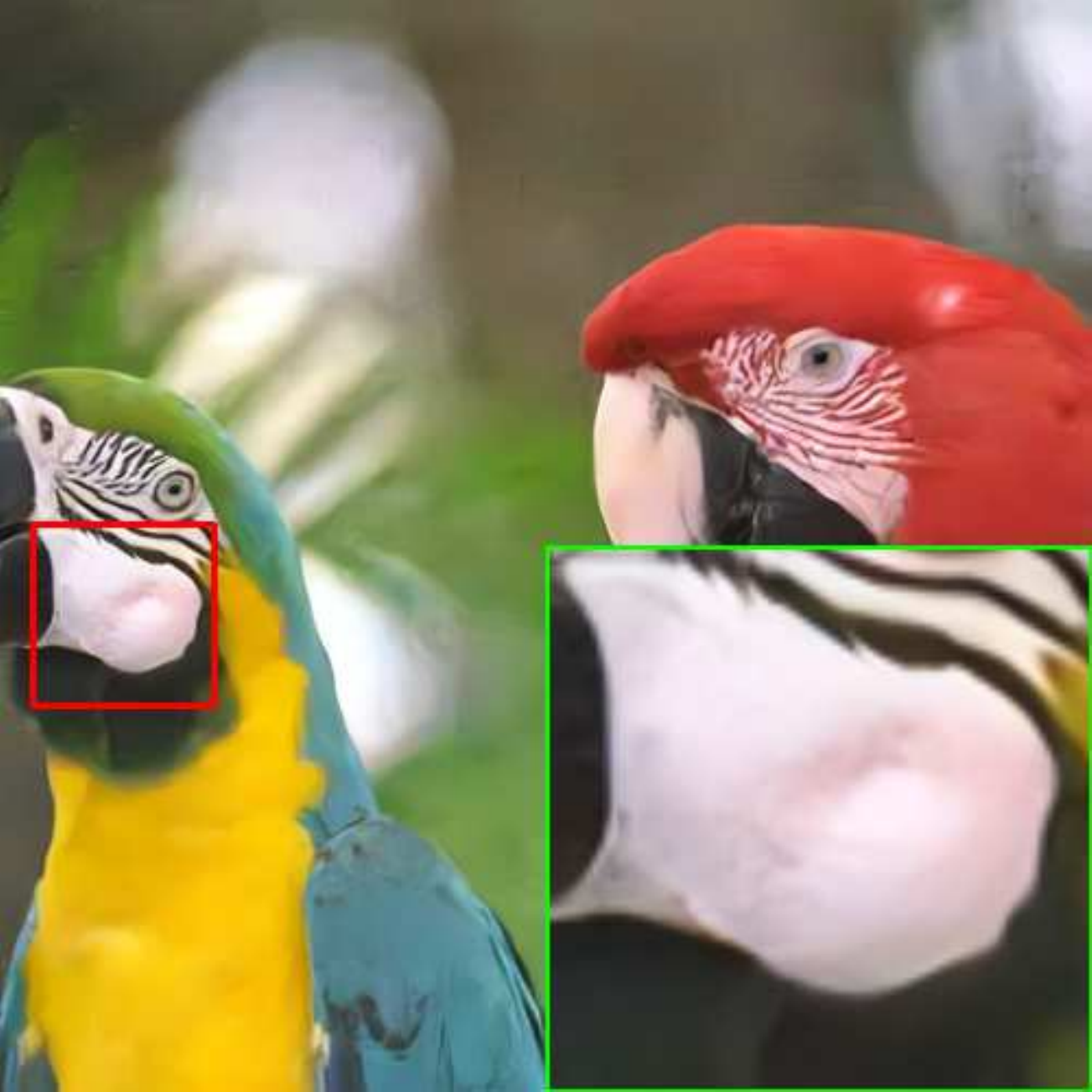}
		\caption{(\uppercase\expandafter{\romannumeral5})}
	\end{subfigure}
    \centering
    \begin{subfigure}{0.15\linewidth}
		\centering
		\includegraphics[width=0.99\linewidth]{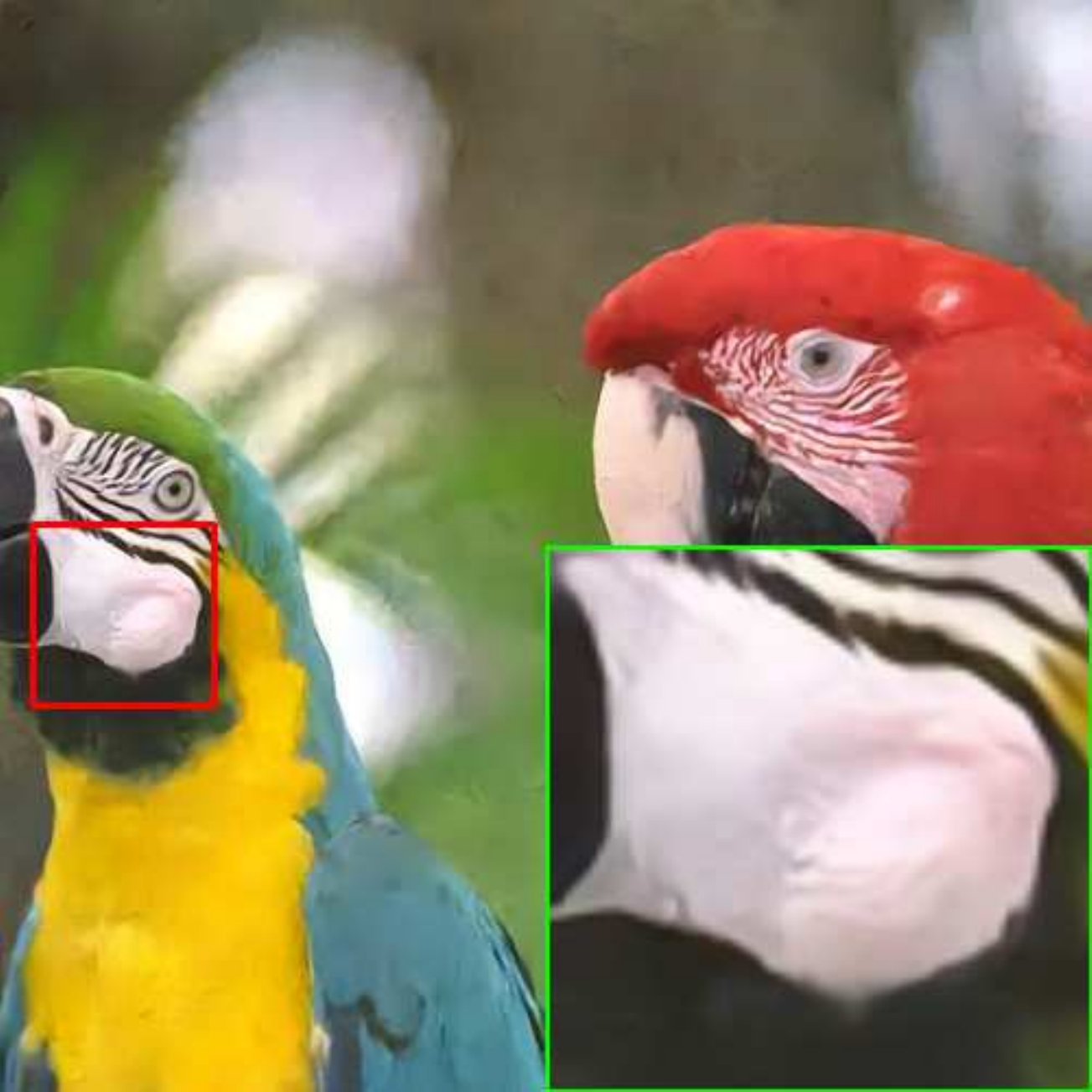}
		\caption{(\uppercase\expandafter{\romannumeral6})}
	\end{subfigure}
    \centering
	\begin{subfigure}{0.15\linewidth}
		\centering
		\includegraphics[width=0.99\linewidth]{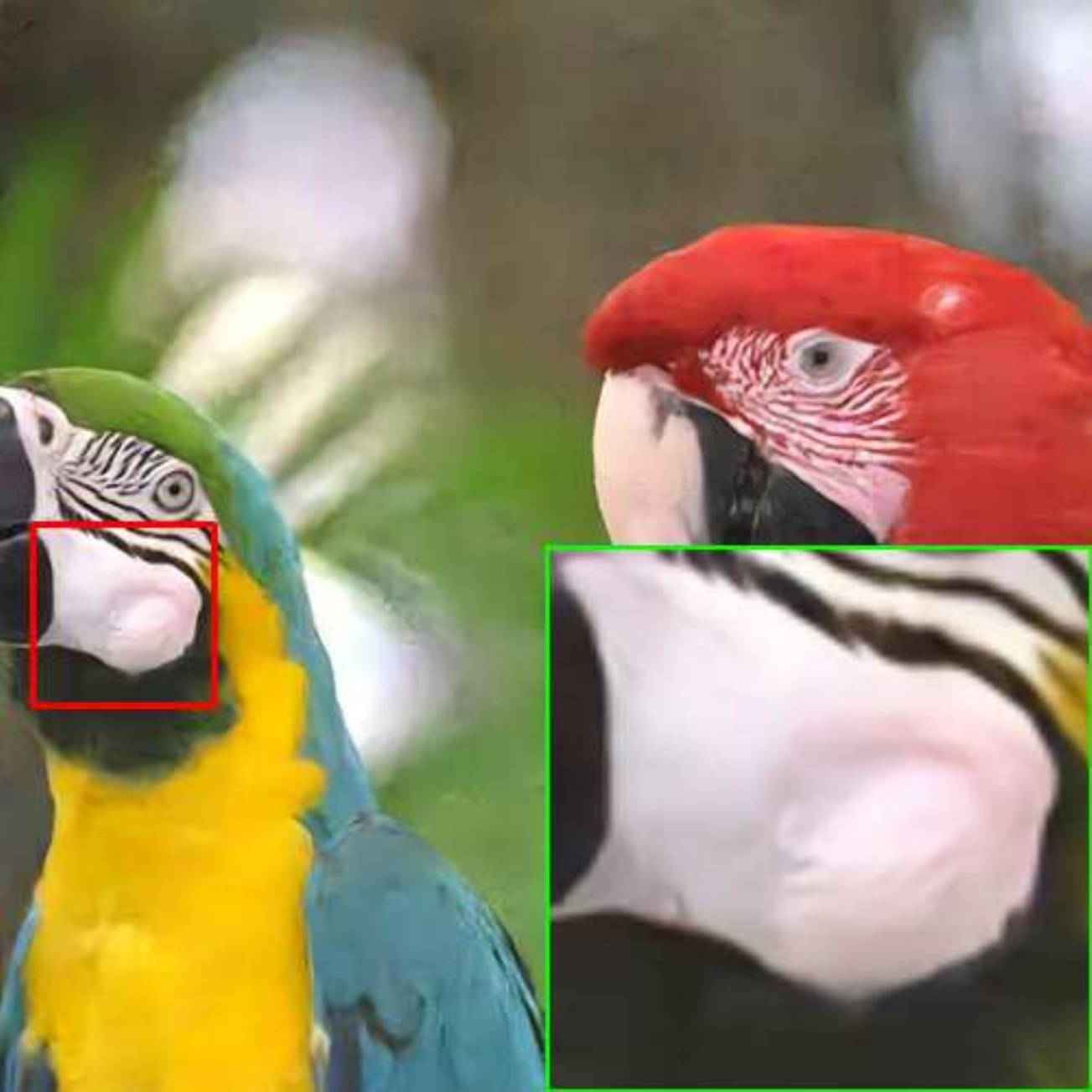}
		\caption{(\uppercase\expandafter{\romannumeral7})}
	\end{subfigure}
    \centering
	\begin{subfigure}{0.15\linewidth}
		\centering
		\includegraphics[width=0.99\linewidth]{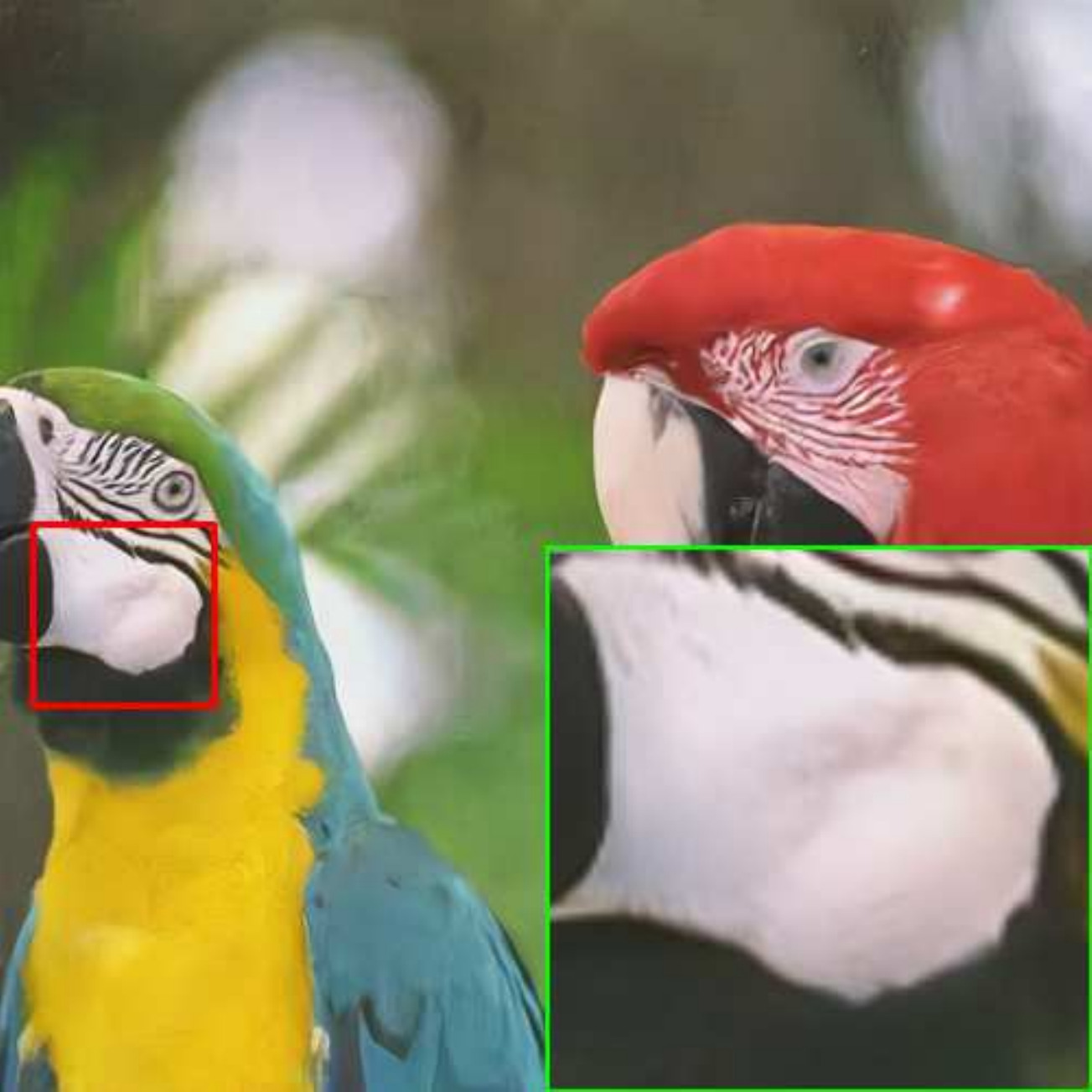}
		\caption{(\uppercase\expandafter{\romannumeral8})}
	\end{subfigure}
    \centering
	\begin{subfigure}{0.15\linewidth}
		\centering
		\includegraphics[width=0.99\linewidth]{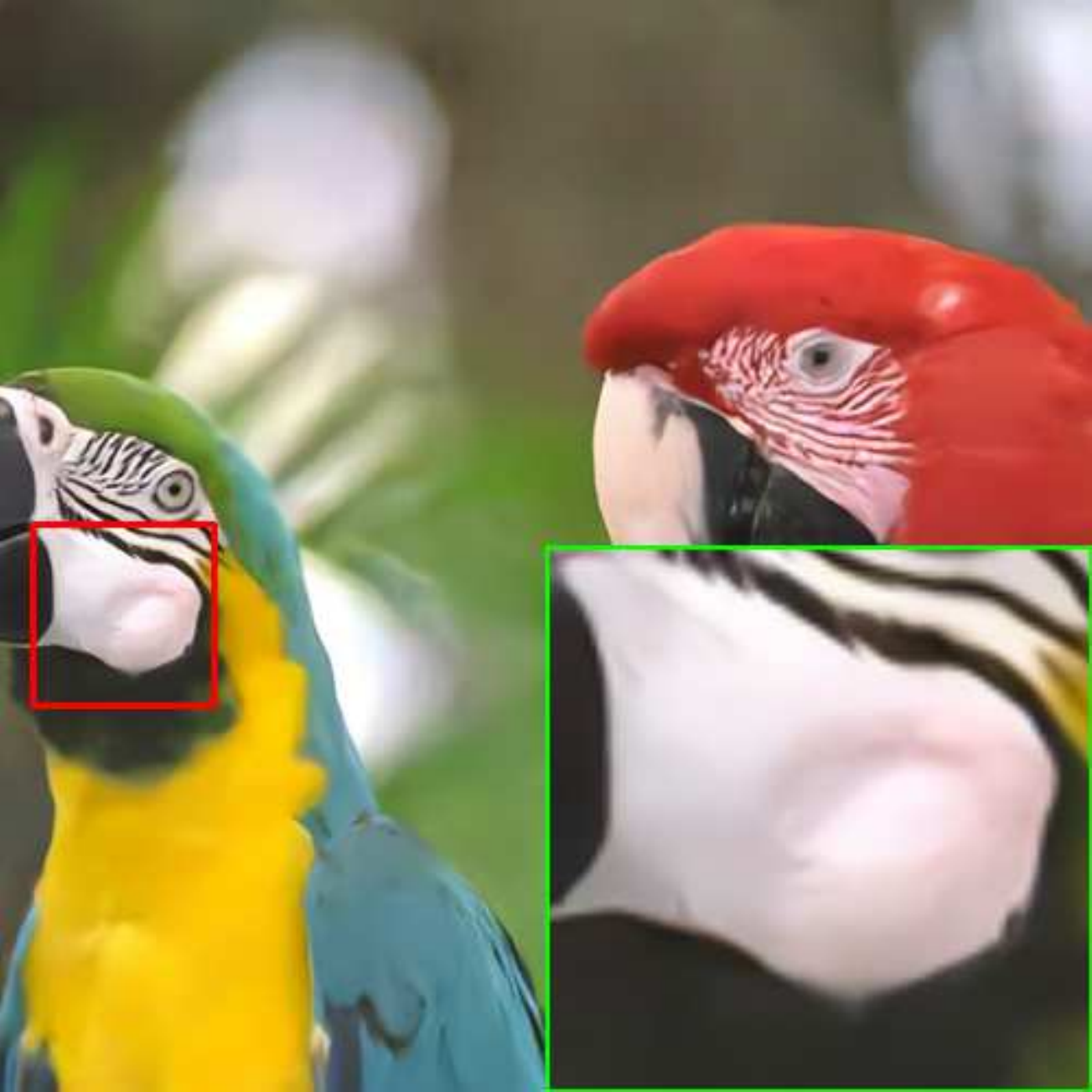}
		\caption{(\uppercase\expandafter{\romannumeral9})}
	\end{subfigure}
    \centering
	\begin{subfigure}{0.15\linewidth}
		\centering
		\includegraphics[width=0.99\linewidth]{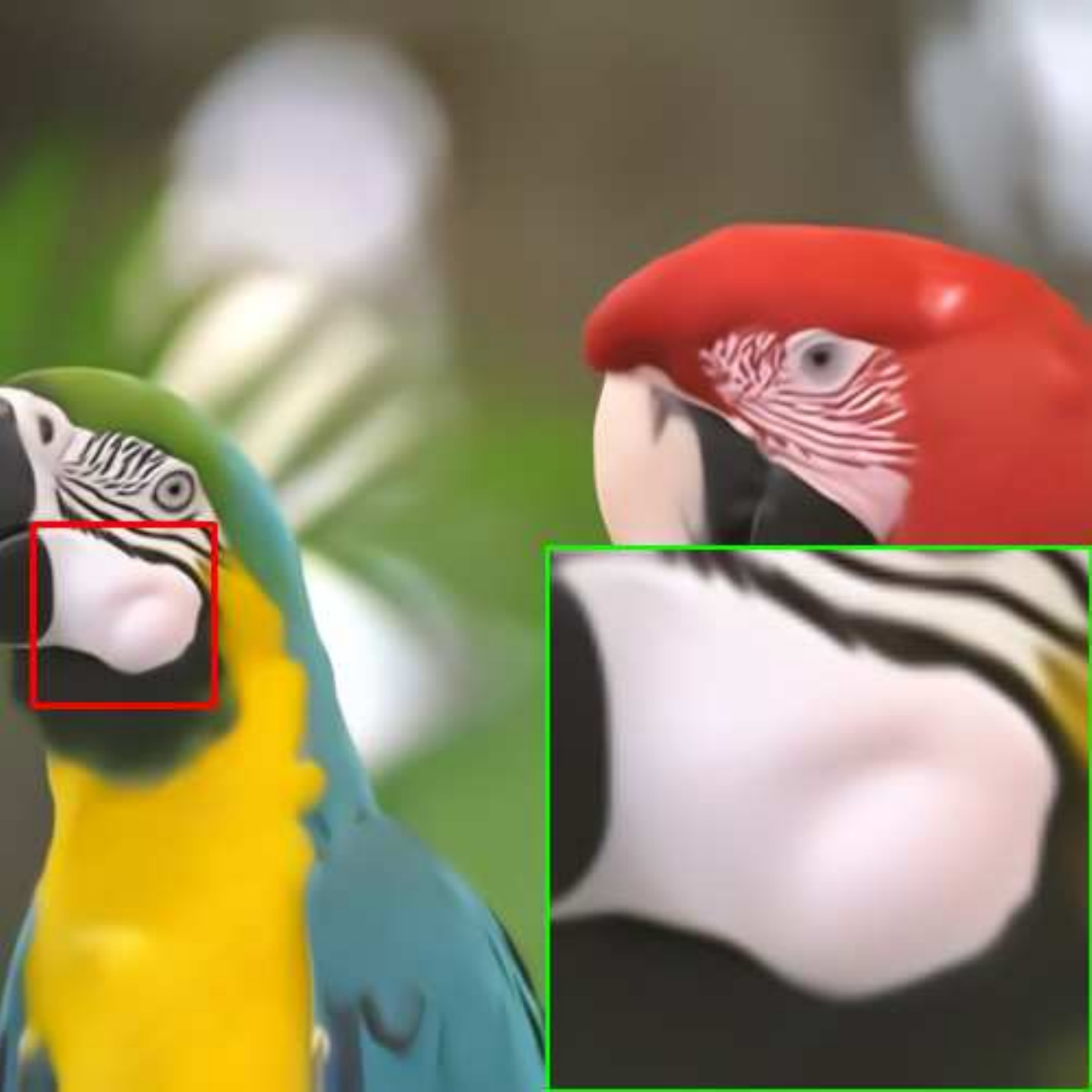}
		\caption{(\uppercase\expandafter{\romannumeral10})}
	\end{subfigure}
    \centering
	\begin{subfigure}{0.15\linewidth}
		\centering
		\includegraphics[width=0.99\linewidth]{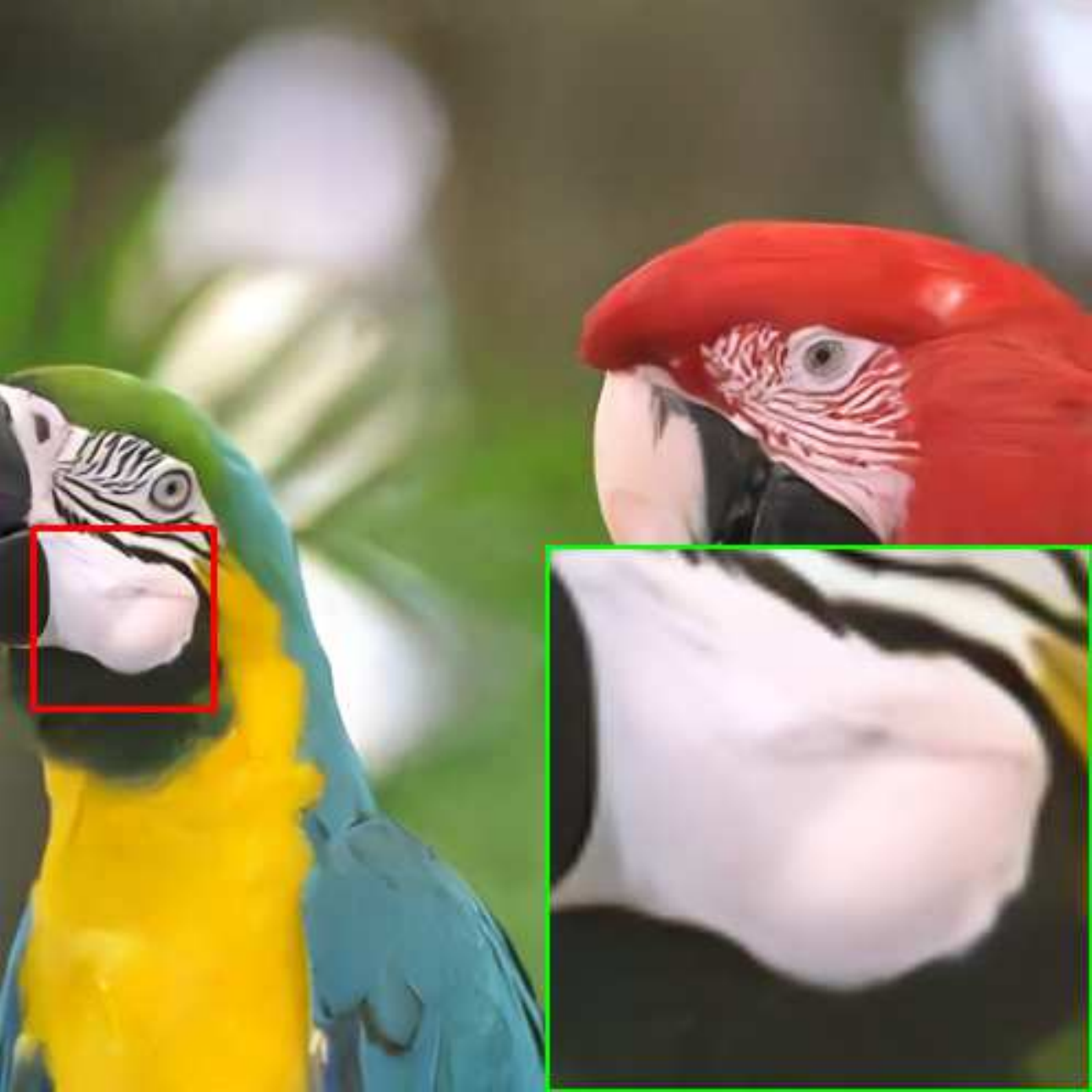}
		\caption{(\uppercase\expandafter{\romannumeral11})}
	\end{subfigure}
    \centering
	\begin{subfigure}{0.15\linewidth}
		\centering
		\includegraphics[width=0.99\linewidth]{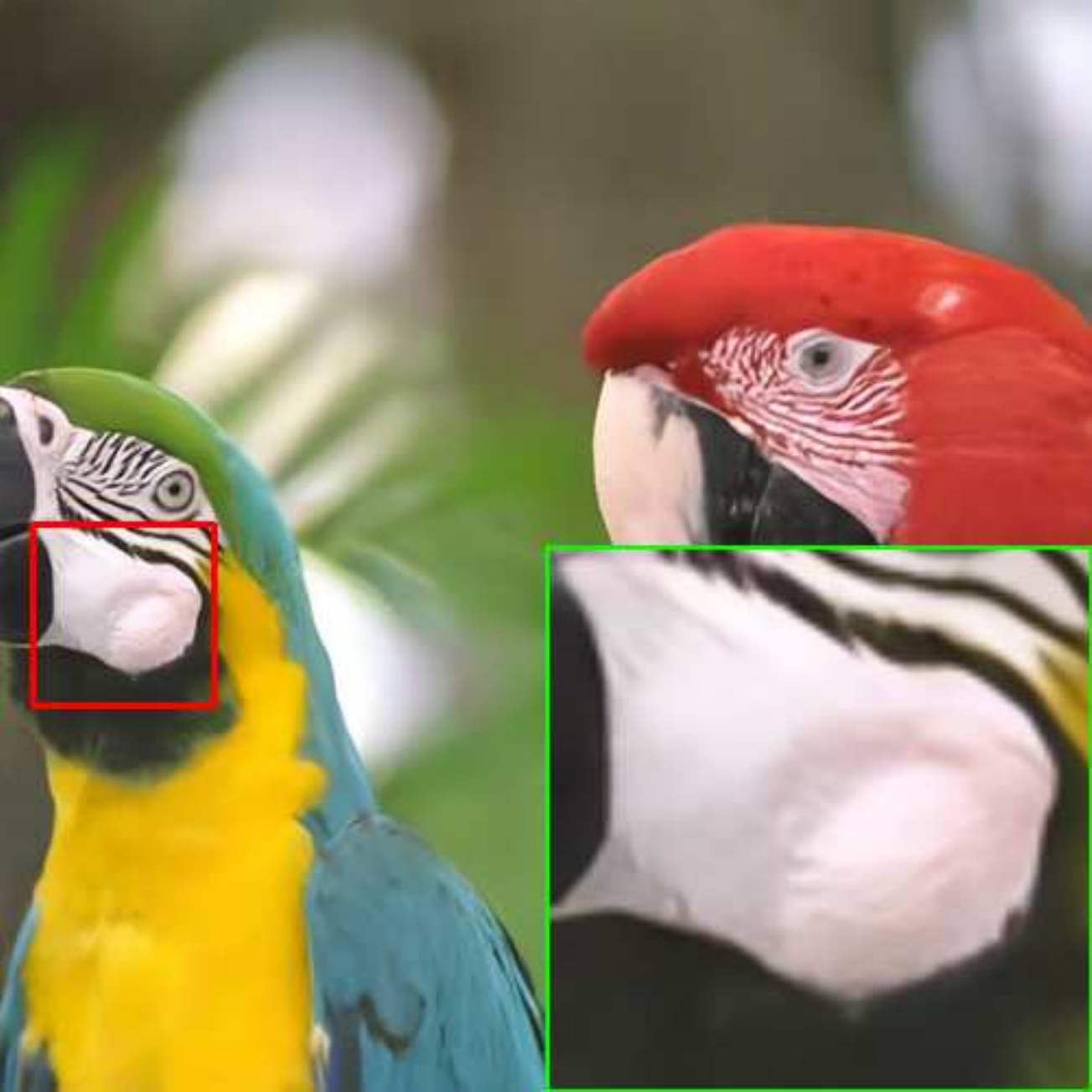}
		\caption{(\uppercase\expandafter{\romannumeral12})}
	\end{subfigure}
\caption{Visual comparison results on the ``kodim23'' image from the Kodak24 dataset at the noise level 50. (\uppercase\expandafter{\romannumeral1}) Ground-truth image, (\uppercase\expandafter{\romannumeral2}) Noisy image / 14.16dB, (\uppercase\expandafter{\romannumeral3}) CBM3D / 30.95dB, (\uppercase\expandafter{\romannumeral4}) MCWNNM / 25.25dB, (\uppercase\expandafter{\romannumeral5}) CDnCNN-S / 31.47dB, (\uppercase\expandafter{\romannumeral6}) CDnCNN-B / 31.12dB, (\uppercase\expandafter{\romannumeral7}) IRCNN / 31.16dB, (\uppercase\expandafter{\romannumeral8}) BUIFD / 29.41dB, (\uppercase\expandafter{\romannumeral9}) FFDNet / 31.50dB, (\uppercase\expandafter{\romannumeral10}) ADNet / 31.47dB, (\uppercase\expandafter{\romannumeral11}) AirNet / 31.50dB, (\uppercase\expandafter{\romannumeral12}) DCANet / 31.86dB.}
\label{fig:kodim23}
\end{figure*}

Fig. \ref{fig:kodim23} presents the visual comparison results on the ``kodim23'' image from the Kodak24 dataset of different denoising models. One can see that our model obtains a visual appealing result, and a favorable trade-off between image detail preservation and noise reduction is obtained.

%Denoising results on the image ``kodim23'' from Kodak24 dataset with noise level of 50.

\subsection{Real noise removal}
For real noise removal, we evaluated the denoising methods on two commonly used public datasets: the SIDD validation set and the DND sRGB images. The two datasets contain real noisy images and near noise-free counterparts, and the counterparts can be provided as the ground truth, which can be utilized to obtain the PSNR and SSIM values.

Table \ref{tab:SIDD_DND} lists the average PSNR and SSIM values of the compared models on the SIDD validation set and the DND sRGB images. One can see that the DCANet model obtains effective denoising performance.

\begin{table*}[htbp]\tiny
\centering
\caption{Quantitative comparison results on the SIDD validation set and the DND sRGB images. The three best results are respectively emphasized in red, blue and green.}
\label{tab:SIDD_DND}
\begin{tabular}{cccccccccc}
\cline{1-10}
Dataset & Metrics & MCWNNM \cite{Xu2017} & TWSC \cite{Xu2018} & DnCNN-B \cite{Zhang2017} & CBDNet \cite{Guo2019} &  RIDNet \cite{Anwar2019} & AINDNet \cite{Kim2020} & VDN \cite{Yue2019} &  DCANet \\
\cline{1-10}
\multirow{3}*{SIDD} & PSNR & 33.40 & 35.33 & 23.66 & 30.78 & 38.71 & \textcolor{green}{38.95} & \textcolor{red}{39.28}	& \textcolor{blue}{39.27}	\\
\cline{2-10}
\multicolumn{1}{c}{} & SSIM & 0.879 & 0.933 & 0.583 & 0.951 & \textcolor{blue}{0.954} & \textcolor{green}{0.952} & \textcolor{red}{0.956}	& \textcolor{red}{0.956} \\
\cline{1-10}
\multirow{3}*{DND} & PSNR & 37.38 & 37.94	& 37.90	& 38.06 & 39.23 & \textcolor{green}{39.37} & \textcolor{blue}{39.38} & \textcolor{red}{39.57} \\
\cline{2-10}
\multicolumn{1}{c}{} & SSIM & 0.929 & 0.940 & 0.943 & 0.942 & \textcolor{red}{0.953} & \textcolor{green}{0.951} & \textcolor{blue}{0.952}	& \textcolor{red}{0.953} \\
\cline{1-10}
\end{tabular}
\end{table*}

\begin{figure*}[htbp]
	\centering
    \captionsetup[subfigure]{labelformat=empty}
	\begin{subfigure}{0.225\linewidth}
		\centering
		\includegraphics[width=0.99\linewidth]{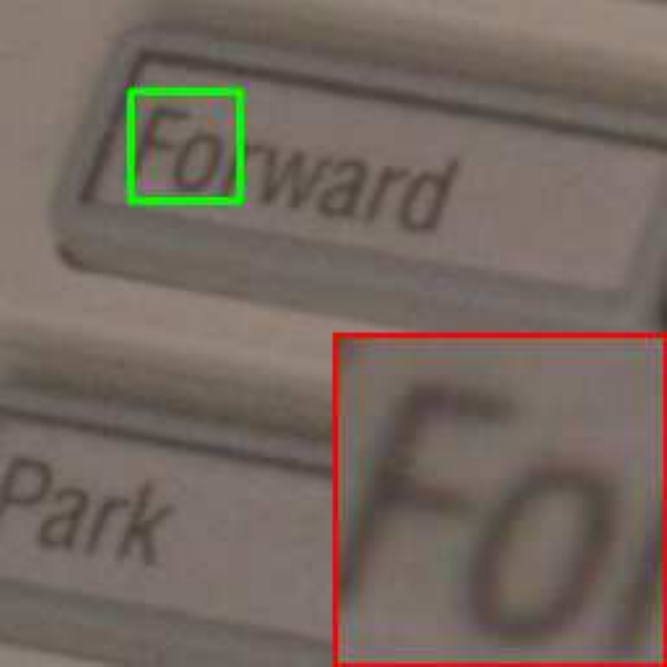}
		\caption{(\uppercase\expandafter{\romannumeral1})}
	\end{subfigure}
    \centering
	\begin{subfigure}{0.225\linewidth}
		\centering
		\includegraphics[width=0.99\linewidth]{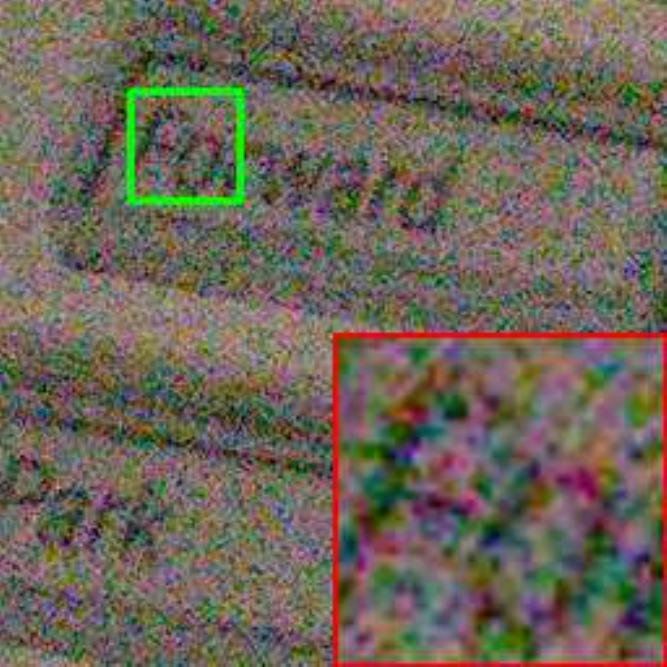}
		\caption{(\uppercase\expandafter{\romannumeral2})}
	\end{subfigure}
    \centering
	\begin{subfigure}{0.225\linewidth}
		\centering
		\includegraphics[width=0.99\linewidth]{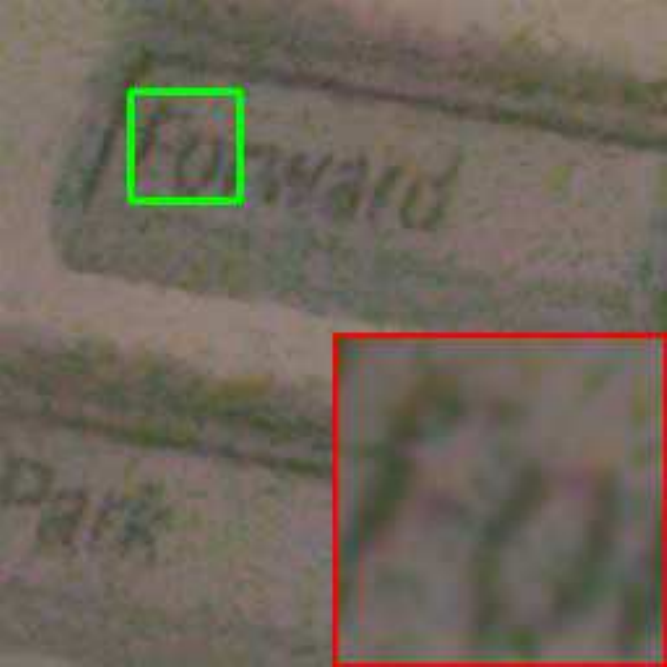}
		\caption{(\uppercase\expandafter{\romannumeral3})}
	\end{subfigure}
    \centering
	\begin{subfigure}{0.225\linewidth}
		\centering
		\includegraphics[width=0.99\linewidth]{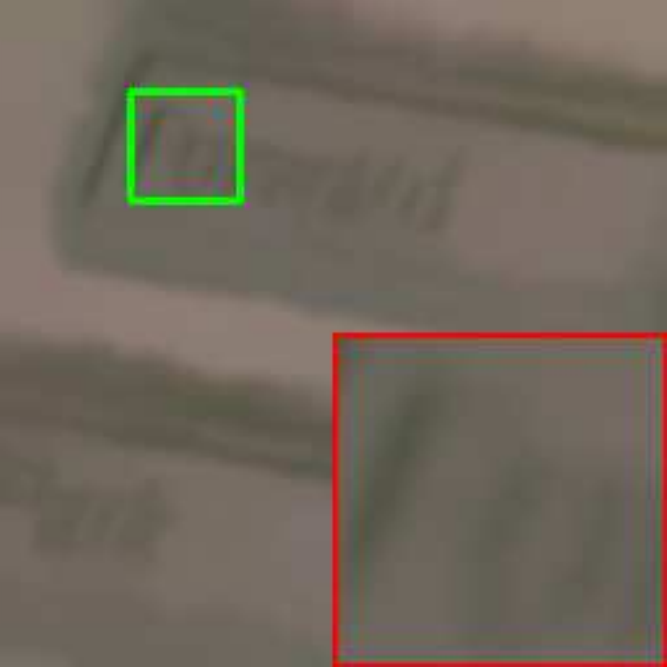}
		\caption{(\uppercase\expandafter{\romannumeral4})}
	\end{subfigure}
    \centering
	\begin{subfigure}{0.225\linewidth}
		\centering
		\includegraphics[width=0.99\linewidth]{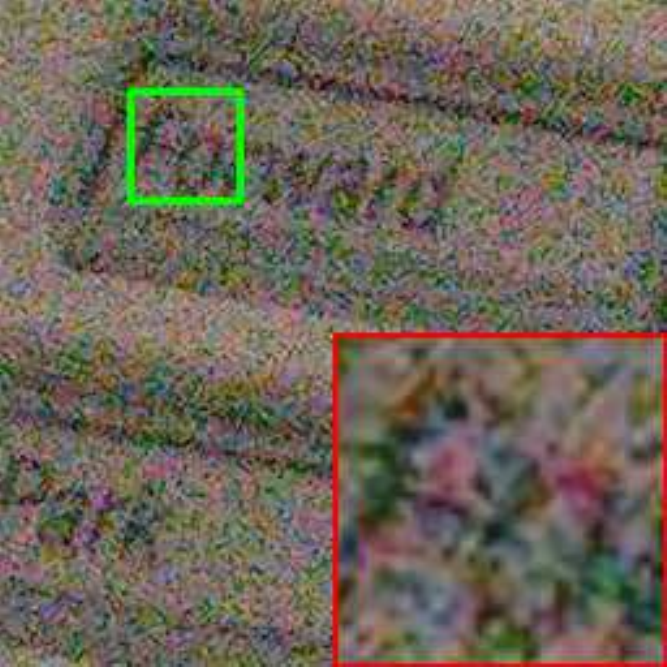}
		\caption{(\uppercase\expandafter{\romannumeral5})}
	\end{subfigure}
    \centering
	\begin{subfigure}{0.225\linewidth}
		\centering
		\includegraphics[width=0.99\linewidth]{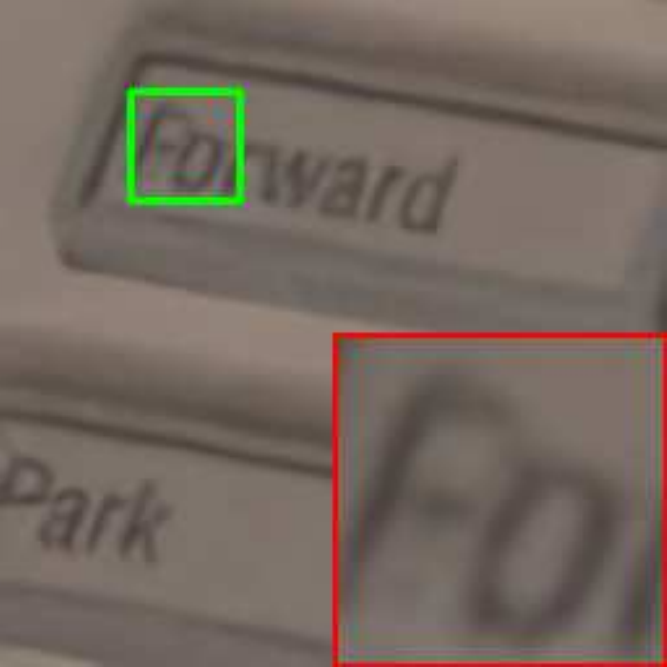}
		\caption{(\uppercase\expandafter{\romannumeral6})}
	\end{subfigure}
    \centering
	\begin{subfigure}{0.225\linewidth}
		\centering
		\includegraphics[width=0.99\linewidth]{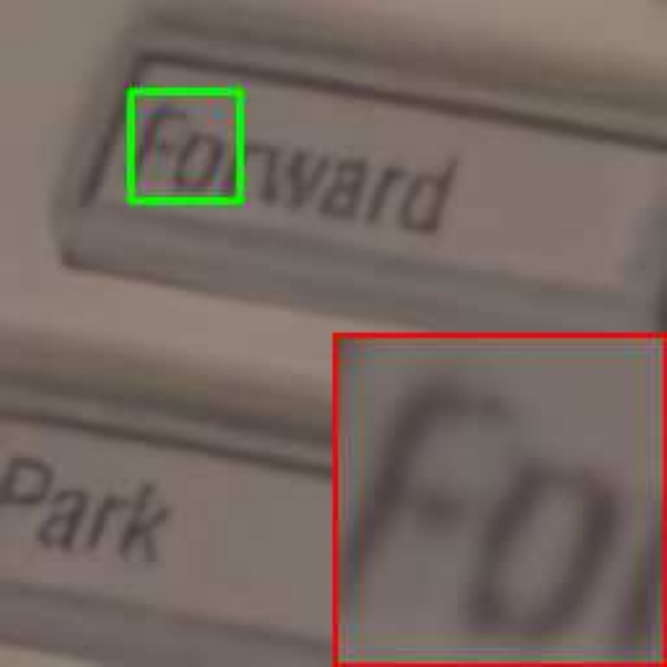}
		\caption{(\uppercase\expandafter{\romannumeral7})}
	\end{subfigure}
    \centering
	\begin{subfigure}{0.225\linewidth}
		\centering
		\includegraphics[width=0.99\linewidth]{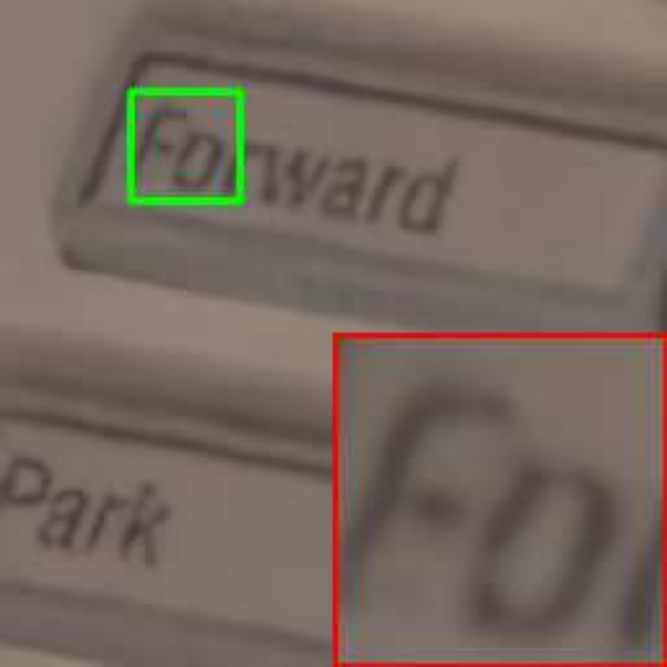}
		\caption{(\uppercase\expandafter{\romannumeral8})}
	\end{subfigure}
\caption{Visual comparison results on the ``11\_4'' image from the SIDD validation set. (\uppercase\expandafter{\romannumeral1}) Ground-truth image / (PSNR/SSIM), (\uppercase\expandafter{\romannumeral2}) Noisy image / (18.25/0.169), (\uppercase\expandafter{\romannumeral3}) MCWNNM / (28.63/0.725), (\uppercase\expandafter{\romannumeral4}) TWSC / (30.42/0.837), (\uppercase\expandafter{\romannumeral5}) CDnCNN-B / (20.76/0.231), (\uppercase\expandafter{\romannumeral6}) AINDNet / (36.24/0.904), (\uppercase\expandafter{\romannumeral7}) VDN / (36.39/0.907), (\uppercase\expandafter{\romannumeral8}) DCANet / (36.36/0.912).}
\label{fig:11_4}
\end{figure*}

Fig. \ref{fig:11_4} shows the visual results of the compared models for real noise removal. It can be found the MCWNNM, TWSC, and CDnCNN-B generated unsatisfactory visual effects. In contrast, the AINDNet, VDN, and the proposed DCANet achieved much better visual quality and PSNR/SSIM values.

%Denoising results on the image ``11_4'' from SIDD testset.

\subsection{Model complexity analysis}
We evaluated the network complexity of the proposed DCANet from the perspectives of the running time, the floating point operations per second (FLOPs), and the number of network parameters. The codes of all the compared denoising models are from the original authors. The BM3D \cite{Dabov2007}, WNNM \cite{Gu2014} and MCWNNM \cite{Xu2017} were run in Matlab (R2020a) environment. The DnCNN-B \cite{Zhang2017}, BUIFD \cite{Helou2020}, IRCNN \cite{ZhangZGZ2017}, FFDNet \cite{Zhang2018}, BRDNet \cite{Tian2020}, CBDNet \cite{Guo2019}, RIDNet \cite{Anwar2019}, DudeNet \cite{Tian2021}, AINDNet \cite{Kim2020}, VDN \cite{Yue2019}, ADNet \cite{TianX2020}, AirNet \cite{Li2022}, and the proposed DCANet were implemented in PyCharm (2021) environment.

Three randomly chosen grayscale and color noisy images of different sizes were used to evaluate the runtime of different denoising models. The noise level was set to 25. We obtained the runtime of every tested method on each image by averaging the time of 20 implementations. It should be noted that our experiments neglected the memory transfer time between the CPU and GPU. Table \ref{tab:time} lists the runtime of the tested denoising methods. It can be found that our DCANet has a fast speed for both grayscale and color images.

%The runtime in noisy images of sizes 256×256, 512×512, 1024×1024.
\begin{table*}[htbp]\scriptsize
\centering
\caption{Runtime (in seconds) comparison of different denoising models on different images.}
\label{tab:time}
\begin{tabular}{cccccccc}
\cline{1-8}
\multirow{3}*{Devices} & \multirow{3}*{Models} &  \multicolumn{2}{c}{$256\times256$}	& \multicolumn{2}{c}{$512\times512$}	& \multicolumn{2}{c}{$1024\times1024$}\\
\cline{3-8}
 \multicolumn{1}{c}{} &  & Gray & Color & Gray & Color & Gray & Color\\
\cline{1-8}
\multirow{5}*{CPU} & BM3D \cite{Dabov2007} & 0.458	& 0.593	& 2.354	& 3.771	& 9.782	& 12.818\\
\cline{2-8}
    & WNNM \cite{Gu2014} & 63.867 & - & 277.003 & - & 1150.842  & - \\
\cline{2-8}
    & MCWNNM \cite{Xu2017} & - & 62.777 & - & 277.623 & - & 1120.112  \\
\cline{1-8}
\multirow{14}*{GPU} & DnCNN-B \cite{Zhang2017} & 0.032	& 0.032	& 0.037	& 0.037	& 0.057	& 0.057\\
\cline{2-8}
    & IRCNN \cite{ZhangZGZ2017}  & 0.030	& 0.030	& 0.030	& 0.030	& 0.030	& 0.030\\
\cline{2-8}
    & BUIFD \cite{Helou2020} & 0.035	& 0.037	& 0.050	& 0.053	& 0.112	& 0.123\\
\cline{2-8}
    & FFDNet \cite{Zhang2018} & 0.031	& 0.030	& 0.031	& 0.030	& 0.032	& 0.030\\
\cline{2-8}
    & AINDNet \cite{Kim2020} & - & 0.531 & - & 2.329 & - & 9.573\\
\cline{2-8}
    & VDN \cite{Yue2019} & 0.144 & 0.162 & 0.607 & 0.597 & 2.367 & 2.376\\
\cline{2-8}
    & ADNet \cite{TianX2020} & 0.031 & 0.033 & 0.035 & 0.045 & 0.051 & 0.093\\
\cline{2-8}
    & AirNet \cite{Li2022}  &  - & 0.143 & - & 0.498 & - & 2.501\\
\cline{2-8}
    & DCANet & 0.058 & 0.059 & 0.099 & 0.104 & 0.252 & 0.258\\
\cline{1-8}
\end{tabular}
\end{table*}

Table \ref{tab:Parameters_FLOPs} reports the number of parameters and FLOPs of different models on grayscale and color images, respectively. It can be observed from Table \ref{tab:Parameters_FLOPs} that some state-of-the-art image blind denoising methods have large numbers of parameters, these models are usually equipped with complex model structures, however the proposed DCANet can obtain competitive or even better performance. For those models have the similar number of parameters, like the BUIFD, BRDNet, DudeNet, and RIDNet, our model achieve better results in most of the tested denoising tasks, especially in color image denoising and real noise removal. Namely, our model obtains a favorable tradeoff between network complexity and denoising performance.

%PSNR results of BSD68 at different noise levels
\begin{table*}[htbp]\scriptsize
\centering
\caption{The number of parameters (in K) and FLOPs (in G) of different models.}
\label{tab:Parameters_FLOPs}
\begin{tabular}{ccccc}
\cline{1-5}
\multirow{3}*{Models} & \multicolumn{2}{c}{Model Parameters} & \multicolumn{2}{c}{FLOPs}\\
\cline{2-5}
 \multicolumn{1}{c}{}  & Gray & Color & $256\times256\times1$ &  $256\times256\times3$\\
\cline{2-5}
DnCNN-B \cite{Zhang2017} & 666 & 668 & 21.93 & 22.12 \\
\cline{1-5}
IRCNN \cite{ZhangZGZ2017} & 186 & 188 & 6.08  & 6.15\\
\cline{1-5}
BUIFD \cite{Helou2020} & 1186 & 1196 & 35.31 & 35.65\\
\cline{1-5}
FFDNet \cite{Zhang2018} & 485 & 852 & 3.97 & 3.14\\
\cline{1-5}
BRDNet \cite{Tian2020} & 1113 & 1117 & 36.48 & 36.63\\
\cline{1-5}
CBDNet \cite{Guo2019} & - & 4365 & - & 20.14\\
\cline{1-5}
RIDNet \cite{Anwar2019} & 1497 & 1499 & 48.90 & 48.98\\
\cline{1-5}
VDN \cite{Yue2019} & 7810 & 7817 & 24.47 & 24.70\\
\cline{1-5}
ADNet \cite{TianX2020} & 519 & 521 & 17.08 & 17.15\\
\cline{1-5}
AINDNet \cite{Kim2020} & - & 13764 & - & -\\
\cline{1-5}
DudeNet \cite{Tian2021} & 1077 & 1079 & 35.35 & 35.43\\
\cline{1-5}
AirNet \cite{Li2022} & - & 8930 & - & 150.64\\
\cline{1-5}
DCANet & 1382 & 1389 & 37.65 & 37.88\\
\cline{1-5}
\end{tabular}
\end{table*}

\section{Conclusion}\label{Conclusion}
In this paper, we propose a dual convolutional neural network with attention (DCANet) for image blind denoising. The proposed DCANet is composed of a noise estimation network, an attention module, and a dual convolutional denoising network. The noise estimation network is applied to estimate the noise in a noisy image. The attention module consists of spatial and channel attention block, and is utilized to filter unimportant information. The dual convolutional denoising network contains two different branches, which not only widens the network to improve its learning ability, but also can capture the complementary image features to enhance the denoising effect. To the best of our knowledge, the combination of the dual CNN and the attention mechanism for image blind denoising has not been investigated before.

Compared with the state-of-the-art denoising models, the proposed DCANet obtains competitive denoising performance. Our proposed DCANet also obtains a favorable trade-off between model complexity and denoising ability, therefore the model can be an option for the practical image denoising tasks. In the future, we will further investigate the noise estimation network to obtain more accurate noise estimation. The application of the DCANet on other low-level visual tasks such as image deraining and super-resolution will also be addressed.

\section*{Declarations}

\bmhead{Funding} The Natural Science Foundation of China No. 61863037, No. 41971392, and the Applied Basic Research Foundation of Yunnan Province under grant No. 202001AT070077.

\bmhead{Authors contribution statement} Wencong Wu conceived and designed the study. Wencong Wu, Guannan Lv, and Yingying Duan performed the experiments. Wencong Wu, and Peng Liang were responsible for drawing figures and tables. Data analysis and collation were carried out by Guannan Lv, and Yingying Duan. Wencong Wu, Yungang Zhang, and Yuelong Xia wrote the paper. Yungang Zhang provided the funding support. Wencong Wu, Yungang Zhang, and Yuelong Xia reviewed and edited the manuscript. All authors read and approved the manuscript.

\bmhead{Ethical and informed consent for data used} The datasets used for this work comply with ethical standards, and the authors were given permission to access these datasets.

\bmhead{Data availability and access} If necessary, data involved in this work can be provided by the corresponding author, and the code of this work is accessible on https://github.com/WenCongWu/DCANet.

\bmhead{Competing Interests} The authors declare that they have no conflicts of interest to this work. The authors declare that they do not have any commercial or associative interest that represents a conflict of interest in connection with the work submitted.

\bibliography{sn-bibliography}

\end{document}